\documentclass{article} 
\usepackage{iclr2024_conference,times}

\usepackage{amsmath,amsfonts,bm}

\def\eqref#1{equation~\ref{#1}}

\def\1{\bm{1}}

\DeclareMathAlphabet{\mathsfit}{\encodingdefault}{\sfdefault}{m}{sl}
\SetMathAlphabet{\mathsfit}{bold}{\encodingdefault}{\sfdefault}{bx}{n}

\usepackage{hyperref}
\usepackage{url}
\usepackage{color}
\usepackage{subfigure}
\usepackage{graphicx}
\usepackage{multirow}
\usepackage{array}
\usepackage{booktabs}
\usepackage{ulem}
\usepackage{enumitem}
\usepackage{algpseudocode}
\usepackage{algorithmicx,algorithm}

\def\blue#1{\textcolor{black}{#1}}

\long\def\comment#1{}

\newtheorem{proposition}{Proposition}

\newtheorem{definition}{Definition}

\makeatletter
\newcommand{\printfnsymbol}[1]{%
  \textsuperscript{\@fnsymbol{#1}}%
}
\makeatother

\makeatletter
\def\@fnsymbol#1{\ensuremath{\ifcase#1\or \dagger\or \ddagger\or
   \mathsection\or \mathparagraph\or \|\or **\or \dagger\dagger
   \or \ddagger\ddagger \else\@ctrerr\fi}}
\makeatother

\title{Inducing high Energy-Latency of Large vision-language Models with \textit{Verbose} Images}

\author{Kuofeng Gao\textsuperscript{\rm 1}, Yang Bai\textsuperscript{\rm 2}, Jindong Gu\textsuperscript{\rm 3}, Shu-Tao Xia\textsuperscript{\rm 1,5}\thanks{Corresponding authors}, Philip Torr\textsuperscript{\rm 3}, Zhifeng Li\textsuperscript{\rm 4}\printfnsymbol{1}, Wei Liu\textsuperscript{\rm 4}\printfnsymbol{1}   \\
\textsuperscript{\rm 1} Tsinghua University \quad \textsuperscript{\rm 2} Tencent Technology (Beijing) Co.Ltd \quad \textsuperscript{\rm 3} University of Oxford \quad \\ \textsuperscript{\rm 4} Tencent Data Platform  \quad  \textsuperscript{\rm 5} Peng Cheng Laboratory \\
\texttt{gkf21@mails.tsinghua.edu.cn, mavisbai@tencent.com}\\
\texttt{\{jindong.gu,philip.torr\}@eng.ox.ac.uk, xiast@sz.tsinghua.edu.cn} \\
\texttt{michaelzfli@tencent.com, wl2223@columbia.edu}
}

\iclrfinalcopy 
\begin{document}

\maketitle

\begin{abstract}
Large vision-language models (VLMs) such as GPT-4 have achieved exceptional performance across various multi-modal tasks. However, the deployment of VLMs necessitates substantial energy consumption and computational resources. Once attackers maliciously induce high energy consumption and latency time (energy-latency cost) during inference  of VLMs, it will exhaust computational resources.
In this paper, we explore this attack surface about availability of VLMs and aim to induce high energy-latency cost during inference of VLMs. We find that high energy-latency cost during inference of VLMs can be manipulated by maximizing the length of generated sequences. To this end, we propose \textbf{\textit{verbose images}}, with the goal of crafting an imperceptible perturbation to induce VLMs to generate long sentences during inference. Concretely, we design three loss objectives. First, a loss is proposed to delay the occurrence of end-of-sequence (EOS) token, where EOS token is a signal for VLMs to stop generating further tokens. Moreover, an uncertainty loss and a token diversity loss are proposed to increase the uncertainty over each generated token and the diversity among all tokens of the whole generated sequence, respectively, which can break output dependency at token-level and sequence-level. 
Furthermore, a temporal weight adjustment algorithm is proposed, which can effectively balance these losses. Extensive experiments demonstrate that our verbose images can increase the length of generated sequences by 7.87$\times$ and 8.56$\times$ compared to original images on MS-COCO and ImageNet datasets, which presents potential challenges for various applications. Our code is available at \url{https://github.com/KuofengGao/Verbose_Images}. 
\end{abstract}

\section{Introduction}
\label{sec:intro}
Large vision-language models (VLMs) \citep{alayrac2022flamingo,chen2022visualgpt,liu2023visual,li2021align,li2023blip}, such as GPT-4 \citep{openai2023gpt4},  have recently achieved remarkable performance in multi-modal tasks, including image captioning, visual question answering, and visual reasoning. However, these VLMs often consist of billions of parameters, necessitating substantial computational resources for deployment. Besides, according to \citet{patterson2021carbon}, both NVIDIA and Amazon Web Services claim that the inference process during deployment accounts for over 90\% of machine learning demand. 

Once attackers maliciously induce high energy consumption and latency time (energy-latency cost) during inference stage, it can exhaust computational resources and reduce availability of VLMs.
The energy consumption is the amount of energy used on a hardware during one inference and latency time is the response time taken for one inference. 
As explored in previous studies, 
sponge samples \citep{shumailov2021sponge} maximize the $\mathcal{L}_2$ norm of activation values across all layers to introduce more representation calculation cost while NICGSlowdown \citep{chen2022nicgslowdown} minimizes the logits of both end-of-sequence (EOS) token and output tokens to induce high energy-latency cost. However, these methods are designed for LLMs or smaller-scale models and cannot be directly applied to VLMs, which will be further discussed in Section~\ref{sec:related}.

\begin{figure*}[t] \centering    
\subfigure[Energy of BLIP-2] { 
\label{BLIP of linear correlation of energy and latency}  
\includegraphics[width=0.23\columnwidth]{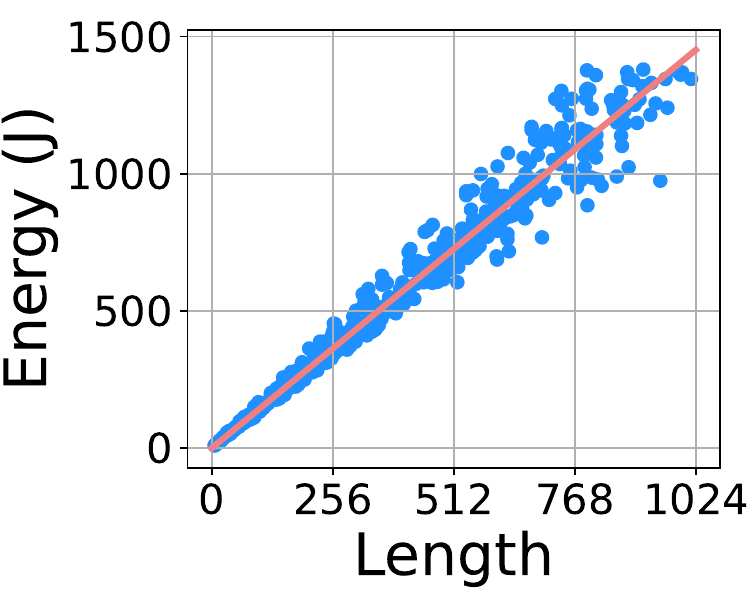}  
}    
\subfigure[Latency of BLIP-2] {
\label{BLIP2 of linear correlation of energy and latency}  \includegraphics[width=0.23\columnwidth]{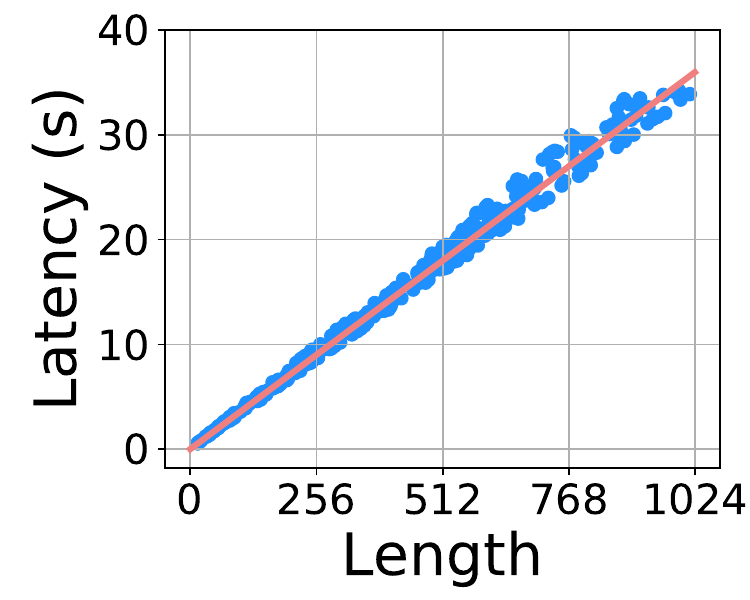}  
}     
\subfigure[Energy of MiniGPT-4] { 
\label{InstructBLIP of linear correlation of energy and latency}  \includegraphics[width=0.23\columnwidth]{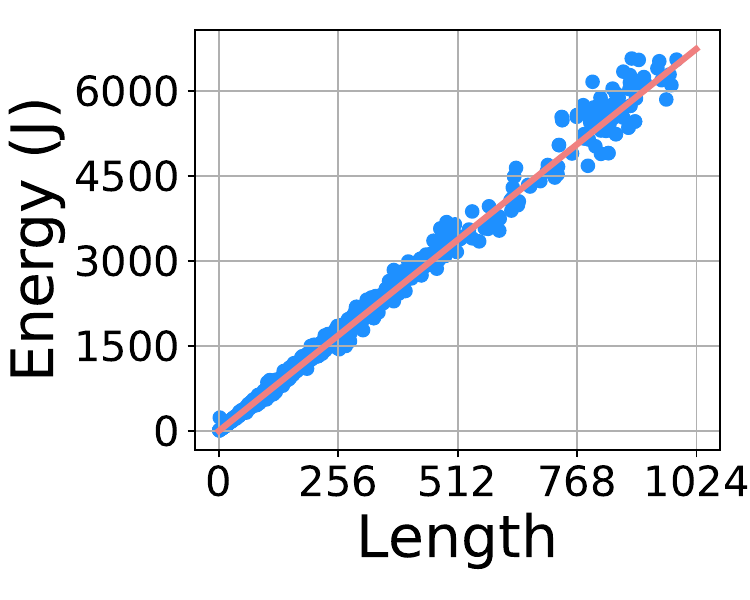}  
}    
\subfigure[Latency of MiniGPT-4] { 
\label{MiniGPT-4 of linear correlation of energy and latency}  \includegraphics[width=0.23\columnwidth]{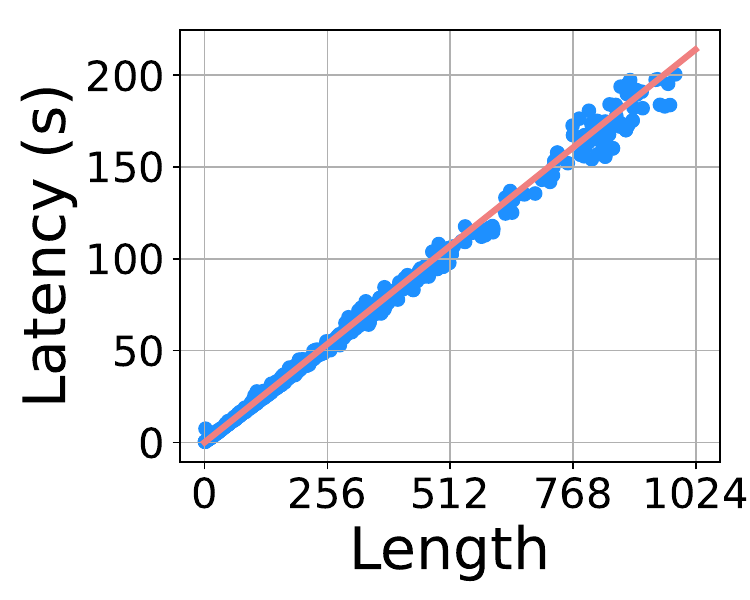}  
}    
\vspace{-1em}
\caption{The approximately positive linear relationship between energy consumption, latency time, and the length of generated sequences in VLMs. Following \citet{shumailov2021sponge}, energy consumption is estimated by NVIDIA Management Library (NVML), and latency time is the response time of an inference. } 
\label{all linear correlation of energy and latency}
\end{figure*}

In this paper, we first conduct a comprehensive investigation on energy consumption, latency time, and the length of generated sequences by VLMs during the inference stage. 
As observed in Fig. \ref{all linear correlation of energy and latency}, both energy consumption and latency time exhibit an approximately positive linear relationship with the length of generated sequences.  
Hence, we can maximize the length of generated sequences to induce high energy-latency cost of VLMs. Moreover, VLMs incorporate the vision modality into impressive LLMs \citep{touvron2023llama,chowdhery2022palm} to enable powerful visual interaction but meanwhile, this integration also introduces vulnerabilities from the manipulation of visual inputs \citep{goodfellow2014explaining}. Consequently, we propose \textbf{\textit{verbose images}} to  craft an imperceptible perturbation to induce VLMs to generate long sentences during inference.


Our objectives for verbose images are designed as follows. (1) \textbf{Delayed EOS loss}: By delaying the placement of the EOS token, VLMs are encouraged to generate more tokens and extend the length of generated sequences. Besides, to accelerate the process of delaying EOS tokens, we propose to break output dependency, following \citet{chen2022nicgslowdown}.  (2) \textbf{Uncertainty Loss}: By introducing more uncertainty over each generated token, it can break the original output dependency at the token level and encourage VLMs to produce more varied outputs and longer sequences. (3) \textbf{Token Diversity Loss}: By promoting token diversity among all tokens of the whole generated sequence,  VLMs are likely to generate a diverse range of tokens in the output sequence, which can break the original output dependency at the sequence level and contribute to longer and more complex sequences. Furthermore, a temporal weight adjustment algorithm is introduced to balance the optimization of these three loss objectives.

In summary, our contribution can be outlined as follows:
\begin{itemize}
\item We conduct a comprehensive investigation and observe that energy consumption and latency time are approximately positively linearly  correlated with the length of generated sequences for VLMs. 

\item We propose \textbf{\textit{verbose images}} to craft an imperceptible perturbation to induce high energy-latency cost for VLMs, which is achieved by delaying the EOS token, enhancing output uncertainty, improving token diversity, and employing a temporal weight adjustment algorithm during the optimization process.

\item Extensive experiments show that our verbose images can increase the length of generated sequences by 7.87$\times$ and 8.56$\times$ relative to original images on MS-COCO and ImageNet across four VLM models. Additionally, our verbose images can produce dispersed attention on visual input and generate complex sequences containing hallucinated contents.
\end{itemize}

\section{Related Work}
\label{sec:related}

\textbf{Large vision-language models (VLMs).} 
Recently, the advanced VLMs, such as BLIP \citep{li2022blip}, BLIP-2 \citep{li2023blip}, InstructBLIP \citep{dai2023instructblip}, and MiniGPT-4 \citep{zhu2023minigpt}, have achieved an enhanced zero-shot performance in various multi-modal tasks. Concretely, BLIP proposes a unified vision and language pre-training framework, while BLIP-2 introduces a query transformer to bridge the modality gap between a vision transformer and an LLM. Additionally, InstructBLIP and MiniGPT-4 both adopt instruction tuning for VLMs to improve the vision-language understanding performance. The integration of the vision modality into VLMs enables visual context-aware interaction, surpassing the capabilities of LLMs. However, this integration also introduces vulnerabilities arising from the manipulation of visual inputs. In our paper, we propose to craft verbose images to induce high energy-latency cost of VLMs.

\textbf{Energy-latency manipulation.} 
The energy-latency manipulation \citep{chen2022nmtsloth,hong2020panda,chen2023dark,liu2023slowlidar} aims to slow down the models by increasing their energy computation and response time during the inference stage, a threat analogous to the denial-of-service (DoS) attacks \citep{pelechrinis2010denial} from the Internet. 
Specifically, \citet{shumailov2021sponge} first observe that a larger representation dimension calculation can introduce more energy-latency cost in LLMs. Hence, they propose to craft sponge samples to maximize the $\mathcal{L}_2$ norm of activation values across all layers, thereby introducing more representation calculation and energy-latency cost.
NICGSlowDown \citep{chen2022nicgslowdown} proposes to increase the number of decoder calls, \textit{i.e.}, the length of the generated sequence, to increase the energy-latency of smaller-scale captioning models. They minimize the logits of both EOS token and output tokens to generate long sentences.

However, these previous methods cannot be directly applied to VLMs for two main reasons. On one hand, they primarily focus on LLMs or smaller-scale models. Sponge samples are designed for LLMs for translations \citep{liu2019roberta} and NICGSlowdown targets for RNNs or LSTMs combined with CNNs for image captioning \citep{anderson2018bottom}. Differently, our verbose images are tailored for VLMs in multi-modal tasks. On the other hand, the objective of NICGSlowdown involves logits of specific output tokens. Nevertheless, current VLMs generate random output sequences for the same input sample, due to advanced sampling policies \citep{holtzman2019curious}, which makes it challenging to optimize objectives with specific output tokens. Therefore, it highlights the need for methods specifically designed for VLMs to induce high energy-latency cost. 

\section{Preliminaries}
\subsection{Threat model}

\textbf{Goals and capabilities.} The goal of our verbose images is to craft an imperceptible image and induce VLMs to generate a sequence as long as possible, thereby increasing the energy consumption and prolonging latency during the victim model's deployment. Specifically, the involved perturbation is restricted within a predefined magnitude in $l_p$ norm, ensuring it difficult to detect. 

\textbf{Knowledge and background.} 
We consider the target VLMs which generate sequences using an auto-regressive process. As suggested in~\citet{bagdasaryan2023ab,qi2023visual}, we assume that the victim VLMs can be accessed in full knowledge, including architectures and parameters.  Additionally, we consider a more challenging scenario where the victim VLMs are inaccessible, as detailed in Appendix \ref{sec: Implementation details} and Appendix \ref{sec: black box}. 


\subsection{Problem formulation}

Consider an image $\bm{x}$, an input text $\bm{c}_{\text{in}}$ and a sequence of generated output tokens $\bm{y}=\{y_1, y_2, ..., y_N\}$, where $y_i$ represents the $i$-th generated token, $N$ is the length of the output sequence and $\bm{c}_{\text{in}}$ is a placeholder $\emptyset$ in image captioning or a question in visual question answering and visual reasoning. 
Based on the probability distribution over generated tokens, VLMs generate one token at one time in an auto-regressive manner.
The probability distribution after the $\operatorname{Softmax}(\cdot)$ layer over the $i$-th generated token can be denoted as $f_i\left(y_1, \cdots, y_{i-1};\ \bm{x};\ \bm{c}_{\text{in}}\right)$. Since we mainly focus on images $\bm{x}$ of VLMs in this paper, we abbreviate it as $f_i\left(\bm{x}\right)$, where $f_i\left(\bm{x}\right) \in \mathbb{R}^{\mathrm{V}}$ and $\mathrm{V}$ is the vocabulary size. Meanwhile, the hidden states across all the layers over the $i$-th generated token are recorded as $g_i\left(y_1, \cdots, y_{i-1};\ \bm{x};\ \bm{c}_{\text{in}}\right)$, abbreviated as  $g_i\left(\bm{x}\right)$, where $g_i\left(\bm{x}\right) \in \mathbb{R}^{\mathrm{C}}$ and $\mathrm{C}$ is the dimension size of hidden states. 

As discussed in Section \ref{sec:intro}, the energy consumption and latency time of an inference are approximately positively linearly related to the length of the generated sequence of VLMs. Hence, we propose to maximize the length $N$ of the output tokens of VLMs by crafting verbose images $\bm{x}'$. To ensure the imperceptibility, we impose an $l_p$ restriction on the imperceptible perturbations, where the perturbation magnitude is denoted as $\epsilon$, such that $||\bm{x}'-\bm{x}||_{p} \le \epsilon$.





\begin{figure*}[t] \centering    
\includegraphics[width=\columnwidth]{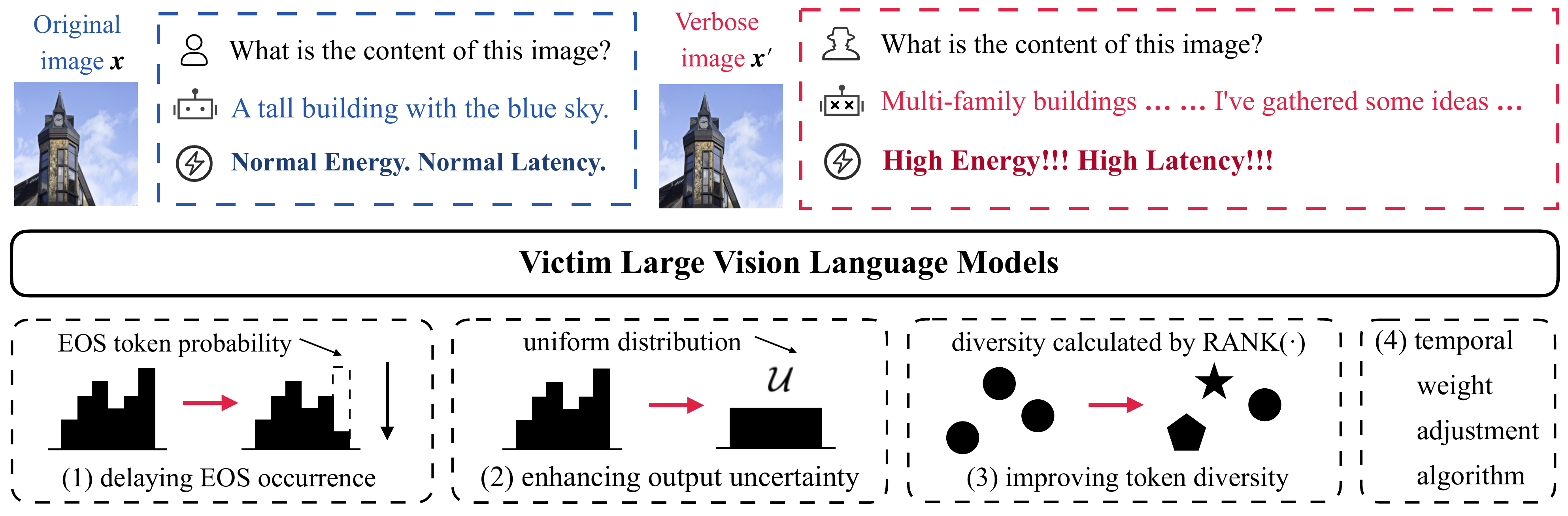}  
\vspace{-2em}
\caption{An overview of verbose images against VLMs to increase the length of generated sequences, thereby inducing higher energy-latency cost. Three losses are designed to craft verbose images by delaying EOS occurrence, enhancing output uncertainty, and improving token diversity. Besides, a temporal weight adjustment algorithm is proposed to better utilize the three objectives.} 
\label{simple illustration}
\vspace{-0.5em}
\end{figure*}

\section{Methodology}
\textbf{Overview.} To increase the length of generated sequences, three loss objectives are proposed to optimize imperceptible perturbations for verbose images in Section \ref{sec:loss design}. Firstly and straightforwardly, we propose a \textbf{delayed EOS loss} to hinder the occurrence of EOS token and thus force the sentence to continue.
However, the auto-regressive textual generation in VLMs establishes an output dependency, which means that the current token is generated based on all previously generated tokens. 
Hence, when previously generated tokens remain unchanged, it is also hard to generate a longer sequence even though the probability of the EOS token has been minimized. 
To this end, we propose to break this output dependency as suggested in \citet{chen2022nicgslowdown}.
Concretely, two loss objectives are proposed at both token-level and sequence-level: a \textbf{token-level uncertainty loss}, which enhances output uncertainty over each generated token, and a \textbf{sequence-level token diversity loss}, which improves the diversity among all tokens of the whole generated sequence. 
Moreover, to balance three loss objectives during the optimization, a temporal weight adjustment algorithm is introduced in Section 
\ref{sec:loss weight}. Fig. \ref{simple illustration} shows an overview of our verbose images.

\subsection{Loss design}

\label{sec:loss design}
\textbf{Delaying EOS occurrence.} For VLMs, the auto-regressive generation process continues until an end-of-sequence (EOS) token is generated or a predefined maximum token length is reached. To increase the length of generated sequences, one straightforward approach is to prevent the occurrence of the EOS token during the prediction process. However, considering that the auto-regressive prediction is a non-deterministic random process, it is challenging to directly determine the exact location of the EOS token occurrence. 
Therefore, we propose to minimize the probability of the EOS token at all positions. This can be achieved through the delayed EOS loss, formulated as:
\begin{equation}
\begin{aligned}
\mathcal{L}_{1}(\bm{x}')=\frac{1}{N} \sum_{i=1}^N f_{i}^{\mathrm{EOS}}\left(\bm{x}'\right),
\end{aligned}
\label{eq:eos loss}
\end{equation}
where $f_{i}^{\mathrm{EOS}}(\cdot)$ is EOS token probability of the probability distribution after the $\operatorname{Softmax}(\cdot)$ layer over the $i$-th generated token.
When reducing the likelihood of every EOS token occurring by minimizing $\mathcal{L}_{1}(\cdot)$, VLMs are encouraged to generate more tokens before reaching the EOS token.


\textbf{Enhancing output uncertainty.} VLMs generate  tokens in the generated sequences based on the generated probability distribution. To encourage predictions that deviate from the order of original generated tokens and focus more on other possible candidate tokens, we propose to enhance output uncertainty over each generated token to facilitate longer and more complex sequences. 
This objective can be implemented by maximizing the entropy of the output probability distribution for each generated token. Based on \citet{shannon1948mathematical}, it can be converted to minimize the Kullback–Leibler ($\mathrm{KL}$) divergence $D_{\mathrm{KL}}(\cdot, \cdot)$ \citep{kullback1951information} between the output probability distribution and a uniform distribution $\mathcal{U}$. 
The uncertainty loss can be formulated as follows:
\begin{equation}
\begin{aligned}
\mathcal{L}_{2}(\bm{x}')=\sum_{i=1}^{N} D_{\mathrm{KL}}\left(f_{i}\left(\bm{x}'\right), \mathcal{U}\right),
\end{aligned}
\label{eq:uncertainty loss}
\end{equation}
where $f_{i}(\cdot)$ is the probability distribution after the $\operatorname{Softmax}(\cdot)$ layer over the $i$-th generated token. The uncertainty loss can introduce more uncertainty in the prediction for each generated token, effectively breaking the original output dependency. Consequently, when the original output dependency is disrupted, VLMs can generate more complex sentences and longer sequences, guided by the delay of the EOS token. 


\textbf{Improving token diversity.} To break original output dependency further, we propose to improve the diversity of hidden states among all generated tokens to explore a wider range of possible outputs. Specifically, the hidden state of a token is the vector representation of a word or subword in VLMs.

\begin{definition}
\label{definition}
Let $\operatorname{Rank}(\cdot)$ indicates the rank of a matrix and $[g_1(\bm{x}');g_2(\bm{x}');\cdots;g_N(\bm{x}')]$ denotes the concatenated matrix of hidden states among all generated tokens. To induce high energy-latency cost, the token diversity is defined as the rank of hidden states among all generated tokens, \textit{i.e.}, $\operatorname{Rank}([g_1(\bm{x}');g_2(\bm{x}');\cdots;g_N(\bm{x}')])$.
\end{definition}

Given by Definition \ref{definition}, increasing the rank of the concatenated matrix of hidden states among all generated tokens yields a more diverse set of hidden states of the tokens. \blue{However, based on \citet{fazel2002matrix}, the optimization of the matrix rank is an NP-hard non-convex problem. } To address this issue, we calculate the nuclear norm of a matrix to approximately measure its rank, as stated in Proposition \ref{proposition2}. Consequently, by denoting the nuclear norm of a matrix as $||\cdot||_{*}$, we can formulate the token diversity loss as follows: 
\begin{equation}
\begin{aligned}
\mathcal{L}_{3}(\bm{x}')=-||[g_1(\bm{x}');g_2(\bm{x}');\cdots;g_N(\bm{x}')]||_{*}.
\end{aligned}
\label{eq:diversity loss}
\end{equation}
This token diversity loss can lead to more diverse and complex sequences, making it hard for VLMs to converge to a coherent output. 
Compared to $\mathcal{L}_{2}(\cdot)$, $\mathcal{L}_{3}(\cdot)$ breaks the original output dependency from diversifying hidden states among all generated tokens. In summary, due to the reduced probability of EOS occurrence by $\mathcal{L}_{1}(\cdot)$, and the disruption of the original output dependency introduced by $\mathcal{L}_{2}(\cdot)$ and $\mathcal{L}_{3}(\cdot)$, our proposed verbose images can induce VLMs to generate a longer sequence and  facilitate a more effective evaluation on the worst-case energy-latency cost of VLMs.

\begin{proposition}
\label{proposition2}
\citep{fazel2002matrix} The rank of the concatenated matrix of hidden states among all generated tokens can be heuristically measured using the nuclear norm of the concatenated matrix of hidden states among all generated tokens.
\end{proposition}

\begin{algorithm}[t]
\caption{Verbose images: Inducing high energy-latency cost of VLMs} 
{\bf Input:} 
Original images $\bm{x}$, the perturbation magnitude $\epsilon$, step size $\alpha$ and optimization iterations $T$. \\
{\bf Output:} Verbose images $\bm{x}'$.
\begin{algorithmic}
\State $\bm{x}_0'$ $\gets$ $\bm{x} + \mathcal{U}(-\epsilon, +\epsilon)$ \Comment{initialize verbose images}
\For{$t \gets 1 \text{ to } T$} \Comment{loop over iterations}
\State $\mathcal{L}_{1}(\bm{x}'_{t-1}), \mathcal{L}_{2}(\bm{x}'_{t-1}), \mathcal{L}_{3}(\bm{x}'_{t-1}) \gets$ Eq. \ref{eq:eos loss}, Eq. \ref{eq:uncertainty loss}, Eq. \ref{eq:diversity loss} \Comment{calculate the losses}
\vspace{0.2em}
\State $\lambda_j'(t) \gets m \times \lambda_j'(t-1) + (1-m) \times \lambda_j(t),\ j=1,2,3$ \Comment{calculate the loss weights}
\State $\bm{x}'_{t} \gets \bm{x}'_{t-1}-\alpha \times \operatorname{sign} (\nabla_{\bm{x}'_{t-1}} 
\sum_{j=1}^3  \lambda_j'(t)  \times \mathcal{L}_{j}(\bm{x}'_{t-1})) $    \Comment{update verbose images}
\State $\bm{x}'_{t} \gets \operatorname{Clip}(\bm{x}'_{t}, -\epsilon, +\epsilon)$ \Comment{clip into $\epsilon$-ball of original images}
\EndFor
\end{algorithmic}
\label{alg_gen_verbose_images}
\end{algorithm}

\subsection{Optimization}
\label{sec:loss weight}
To combine the three loss functions, $\mathcal{L}_{1}(\cdot)$, $\mathcal{L}_{2}(\cdot)$, and $\mathcal{L}_{3}(\cdot)$ into an overall objective function, we propose to assign three weights $\lambda_1$, $\lambda_2$, and $\lambda_3$ to the $\mathcal{L}_{1}(\cdot)$, $\mathcal{L}_{2}(\cdot)$, and $\mathcal{L}_{3}(\cdot)$ and sum them up to obtain the final objective function as follows: 
\begin{equation}
\begin{aligned}
\min_{\bm{x}'}\ \lambda_1 \times \mathcal{L}_{1}(\bm{x}') + \lambda_2 \times \mathcal{L}_{2}(\bm{x}') + \lambda_3 \times \mathcal{L}_{3}(\bm{x}'),\quad s.t.\ ||\bm{x}'-\bm{x}||_{p} \le \epsilon,
\end{aligned}
\label{eq:three losses of final objective}
\end{equation}
where $\epsilon$ is the perturbation magnitude to ensure the imperceptibility. To optimize this objective, we adopt the projected gradient descent (PGD) algorithm, as proposed by  \citet{madry2017towards}.  
PGD algorithm is an iterative optimization technique that updates the solution by taking steps in the direction of the negative gradient while projecting the result back onto the feasible set. We denote verbose images at the $t$-th step as $\bm{x}'_t$ and the gradient descent step is as follows:
\begin{equation}
\begin{aligned}
\bm{x}'_{t}=\bm{x}'_{t-1}-\alpha \times \operatorname{sign} ( \nabla_{\bm{x}'_{t-1}} & (\lambda_1  \times \mathcal{L}_{1}(\bm{x}'_{t-1}) + \lambda_2 \times \mathcal{L}_{2}(\bm{x}'_{t-1}) + \lambda_3 \times \mathcal{L}_{3}(\bm{x}'_{t-1})) ),\\ & s.t.\ ||\bm{x}'_{t}-\bm{x}||_{p} \le \epsilon,
\end{aligned}
\label{eq:pgd to optimization}
\end{equation}
where $\alpha$ is the step size. Since  different loss functions have different convergence rates during the iterative optimization process, we propose a \textbf{\textit{temporal weight adjustment algorithm}} to achieve a better balance among these three loss objectives. 
Specifically, we incorporate normalization scaling and temporal decay functions, $\mathcal{T}_1(t)$, $\mathcal{T}_2(t)$, and $\mathcal{T}_3(t)$, into the optimization weights $\lambda_1(t)$, $\lambda_2(t)$, and $\lambda_3(t)$ of $\mathcal{L}_{1}(\cdot)$, $\mathcal{L}_{2}(\cdot)$, and $\mathcal{L}_{3}(\cdot)$. It can be formulated as follows: 
\begin{equation}
\begin{aligned}
&\lambda_1(t) = ||\mathcal{L}_{2}(\bm{x}'_{t-1})||_1\ /\ ||\mathcal{L}_{1}(\bm{x}'_{t-1})||_1\ /\ \mathcal{T}_1(t), \\
&\lambda_2(t) = ||\mathcal{L}_{2}(\bm{x}'_{t-1})||_1\ /\ ||\mathcal{L}_{2}(\bm{x}'_{t-1})||_1\ /\ \mathcal{T}_2(t),\\
&\lambda_3(t) = ||\mathcal{L}_{2}(\bm{x}'_{t-1})||_1\ /\ ||\mathcal{L}_{3}(\bm{x}'_{t-1})||_1\ /\ \mathcal{T}_3(t),
\end{aligned}
\label{eq:dynamic weight of temporal decay}
\end{equation}
where the temporal decay functions are set as:
\begin{equation}
\begin{aligned}
\mathcal{T}_1(t)=a_1 \times \operatorname{ln}(t) + b_1,\  \mathcal{T}_2(t)=a_2 \times \operatorname{ln}(t) + b_2, \ \mathcal{T}_3(t)=a_3 \times \operatorname{ln}(t) + b_3.
\end{aligned}
\label{eq:temporal decay function}
\end{equation}
Besides, a momentum value $m$ is introduced into the update process of weights. This involves taking into account not only  current weights but also  previous weights when updating losses, which helps smooth out the weight updates. The algorithm of our verbose images is summarized in Algorithm \ref{alg_gen_verbose_images}.


\section{Experiments}

\subsection{Experimental Setups}
\textbf{Models and datasets.} We consider four open-source and advanced large vision-language models as our evaluation benchmark, including BLIP \citep{li2022blip}, BLIP-2 \citep{li2023blip}, InstructBLIP \citep{dai2023instructblip}, and MiniGPT-4 \citep{zhu2023minigpt}. 
Concretely, we adopt the BLIP with the basic multi-modal mixture of encoder-decoder model in 224M version, BLIP-2 with an OPT-2.7B LM \citep{zhang2022opt}, InstructBLIP and MiniGPT-4 with a Vicuna-7B LM \citep{chiang2023vicuna}. These models perform the captioning task for the image under their default prompt template. Results of more tasks are in Appendix \ref{sec: More tasks}.
We randomly choose the 1,000 images from MS-COCO \citep{lin2014microsoft} and ImageNet \citep{deng2009imagenet} dataset, respectively, as our evaluation dataset. More details about target models are shown in Appendix \ref{sec: Target models}.

\textbf{Baselines and setups.}
For evaluation, we consider original images, images with random noise, sponge samples, and NICGSlowDown as baselines. For sponge samples, NICGSlowDown, and our verbose images, we perform the projected gradient descent (PGD) \citep{madry2017towards} algorithm in $T=1,000$ iterations. Besides, in order to ensure the imperceptibility, the perturbation magnitude  is set as 
$\epsilon=8$ within $l_{\infty}$ restriction, following \citet{carlini2019evaluating}, and the step size is set as $\alpha=1$. The default maximum length of generated sequences of VLMs is set as $512$ and the sampling policy  is configured to use nucleus sampling \citep{holtzman2019curious}. For our verbose images, the parameters of loss weights are $a_1=10$, $b_1=-20$, $a_2=0$, $b_2=0$, $a_3=0.5$, and $b_3=1$ and the momentum of our optimization is $m=0.9$. More details about setups are listed in Appendix \ref{sec: Experimental setups}. 


\textbf{Evaluation metrics.} We calculate the energy consumption (J) and the latency time (s) during  inference on one single GPU. Following \citet{shumailov2021sponge}, the energy consumption and latency time are measured by the NVIDIA Management Library (NVML) and the response time cost of an inference, respectively. Besides, the length of generated sequences is also regarded as a metric. Considering the randomness of sampling modes in VLMs, we report the average evaluation results run over three times.

\begin{table*}[]
\caption{The length of generated sequences, energy consumption (J), and latency time (s) of five categories of visual images against four VLM models, including BLIP, BLIP-2, InstructBLIP, and MiniGPT-4, on two datasets, namely MS-COCO and ImageNet. The best results are marked in \textbf{bold}.}
\label{tab:main results}
\centering
\small
\setlength\tabcolsep{6.75pt}{
\begin{tabular}{@{}ll|ccc|ccc@{}}
\toprule
\multirow{2}{*}{VLM model} & \multicolumn{1}{l|}{\multirow{2}{*}{Method}} & \multicolumn{3}{c|}{MS-COCO} & \multicolumn{3}{c}{ImageNet} \\ 
 & & Length & Latency & Energy & Length & Latency & Energy  \\
\midrule 
\multirow{5}{*}{BLIP} & Original & 10.03 &  0.21 & 9.51 & 10.17 & 0.22 & 9.10 \\
& Noise & 9.98 & 0.17 & 8.57 & 9.87 & 0.18 & 8.29 \\
& Sponge samples & 65.83 & 1.10 & 73.57 & 76.67 &  1.26 & 86.00 \\
& NICGSlowDown & 179.42 & 2.84 & 220.73 & 193.68 & 2.98 & 243.84 \\
& \textbf{Verbose images (Ours)} &  \textbf{318.66} & \textbf{5.13} & \textbf{406.65} & \textbf{268.25} & \textbf{4.31} & \textbf{344.91} \\
\midrule
\multirow{5}{*}{BLIP-2} & Original & 8.82 & 0.39 & 16.08 & 8.11 & 0.37 & 15.39 \\
& Noise & 9.55 & 0.43 & 17.53 & 8.37 & 0.44 & 19.39 \\
& Sponge samples & 22.53 & 0.73 & 30.20 & 43.59 & 1.51 & 63.27 \\
& NICGSlowDown & 103.54 & 3.78 & 156.61 & 129.68 & 4.34 & 180.06 \\
& \textbf{Verbose images (Ours)} & \textbf{226.72} & \textbf{7.97} & \textbf{321.59} & \textbf{250.72} & \textbf{10.26} & \textbf{398.58} \\
\midrule
\multirow{5}{*}{InstructBLIP} & Original & 63.79 & 2.97 & 151.80 & 54.40 & 2.60 & 128.03 \\
& Noise & 62.76 & 2.91 & 148.64 &  53.01 & 2.50 & 125.42 \\
& Sponge samples & 92.69 & 4.10 & 209.81 & 80.26 & 3.55 & 175.17 \\
& NICGSlowDown & 93.70 & 4.08 &  200.51 & 81.64 & 3.56 & 174.44 \\
& \textbf{Verbose images (Ours)} & \textbf{140.35} & \textbf{6.15} & \textbf{316.06} & \textbf{131.79}  & \textbf{6.05} & \textbf{300.43} \\
\midrule
\multirow{5}{*}{MiniGPT-4} & Original &  45.29 & 10.39 &  329.50 & 40.93 &  9.11 & 294.68   \\
& Noise & 45.15 & 10.35 & 327.04 & 47.78 & 10.98 & 348.66 \\
& Sponge samples & 220.30 & 43.84 & 1390.73 & 228.70 & 47.74 & 1528.58 \\
& NICGSlowDown & 232.80 & 46.39 & 1478.74 & 245.51 &  51.22 & 1624.06 \\
& \textbf{Verbose images (Ours)} & \textbf{321.35} & \textbf{67.14} & \textbf{2113.29} & \textbf{321.24} & \textbf{64.31} &  \textbf{2024.62} \\
\bottomrule
\end{tabular}}
\vspace{-2em}
\end{table*}


\subsection{Main Results}
Table \ref{tab:main results} compares the length of generated sequences, energy consumption, and latency time of original images, images with random noise, sponge samples, NICGSlowdown, and our verbose images. 
The original images serve as a baseline, providing reference values for comparison. 
When random noise is added to the images, the generated sequences exhibit a similar length to those of the original images. It illustrates that it is necessary to optimize a handcrafted perturbation to induce high energy-latency cost of VLMs. The sponge samples and NICGSlowdown can generate longer sequences compared to original images. However, the increase in length is still smaller than that of our verbose images. This can be attributed to the reason that the additional computation cost introduced by sponge samples and the objective for longer sequences in smaller-scale models introduced by NICGSlowdown cannot directly be transferred to induce high energy-latency cost for VLMs.

\begin{figure*}[t] \centering    
\subfigure[BLIP] { 
\label{BLIP of length distribution}  
\includegraphics[width=0.232\columnwidth]{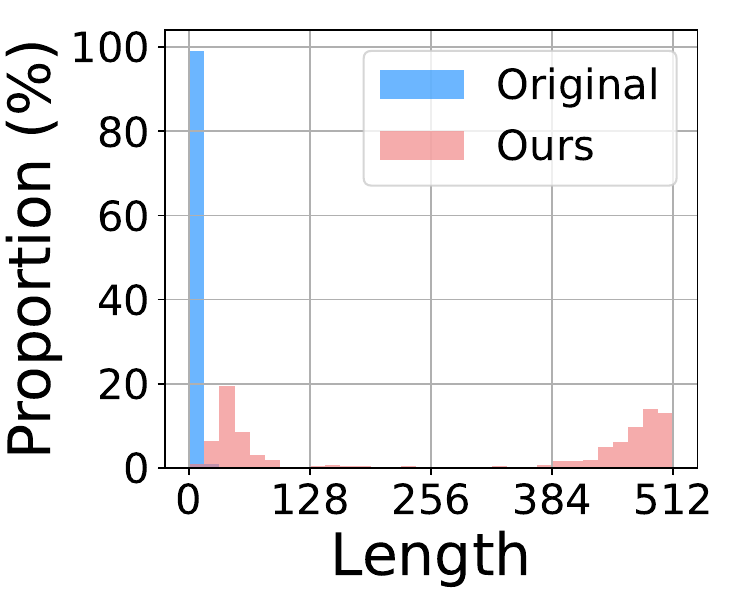}  
}    
\subfigure[BLIP-2] {
\label{BLIP2 of length distribution}  \includegraphics[width=0.232\columnwidth]{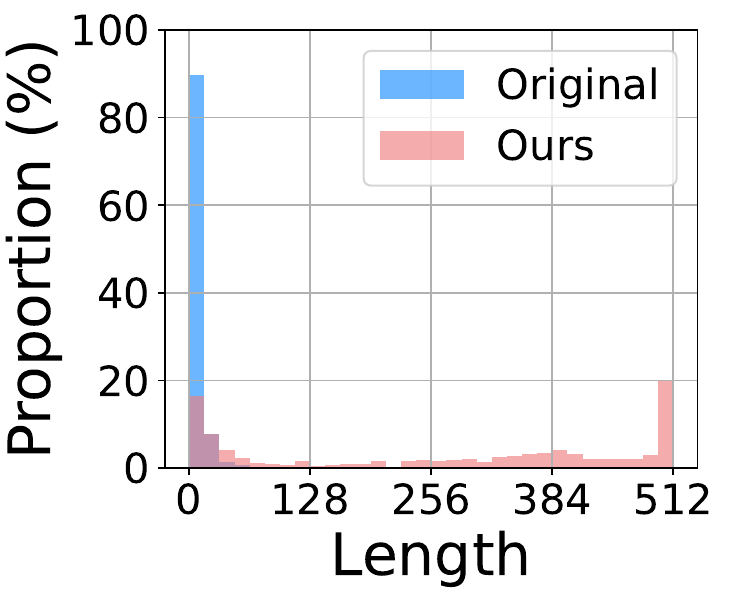}  
}     
\subfigure[InstructBLIP] { 
\label{InstructBLIP of length distribution}  \includegraphics[width=0.232\columnwidth]{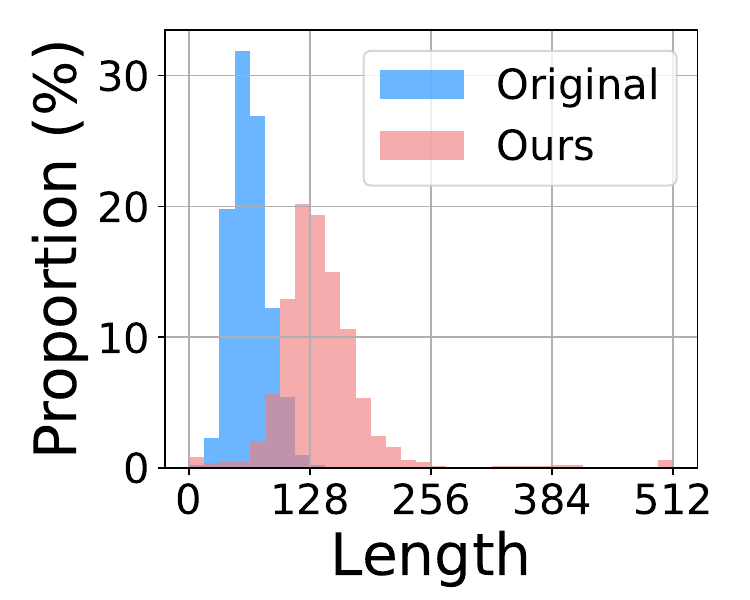}  
}    
\subfigure[MiniGPT-4] { 
\label{MiniGPT-4 of length distribution}  \includegraphics[width=0.232\columnwidth]{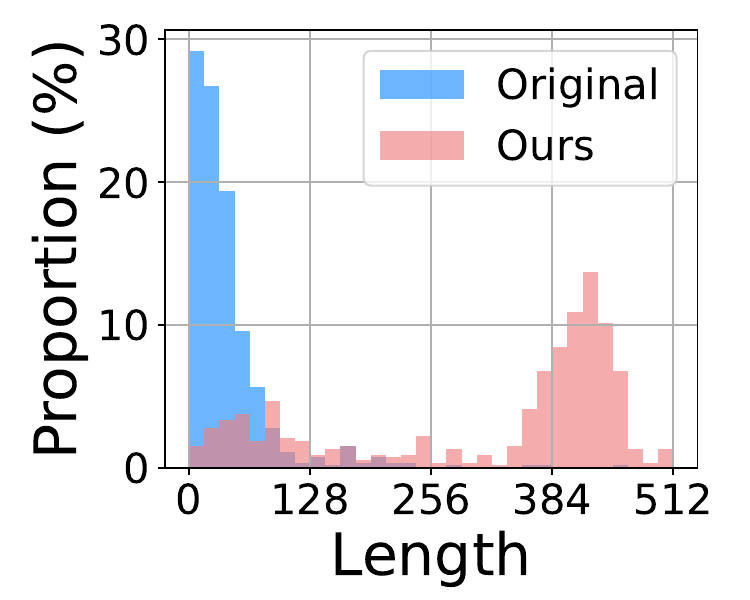}  
}    
\vspace{-1em}
\caption{The length distribution of four VLM models: (a) BLIP. (b) BLIP-2. (c) InstructBLIP. (d) MiniGPT-4. The peak of length distribution of our verbose images shifts towards longer sequences.} 
\vspace{-1em}

\label{all length distribution}
\end{figure*}

Our verbose images can  increase the length of generated sequences and introduce the highest energy-latency cost among all these methods. Specifically, our verbose images can increase the average length of generated sequences by 7.87$\times$ and 8.56$\times$ relative to original images on the MS-COCO and ImageNet datasets, respectively. These results demonstrate the superiority of our verbose images.
In addition, we visualize the length distribution of output sequences generated by four VLMs on original images and our verbose images in Fig. \ref{all length distribution}. Compared to original images, the distribution peak for sequences generated using our verbose images exhibits a shift towards the direction of the longer length, confirming the effectiveness of our verbose images in generating longer sequences. We conjecture that the different shift magnitudes are due to different architectures, different training policies, and different parameter quantities in these VLMs. More results of length distribution are shown in Appendix \ref{sec: More results of length distribution}.



\subsection{Discussions}
To better reveal the mechanisms behind our verbose images, we conduct two further studies, including the visual interpretation where we adopt Grad-CAM \citep{selvaraju2017grad} to generate the attention maps and the textual interpretation where we evaluate the object hallucination in generated sequences by CHAIR \citep{rohrbach2018object}. 

\begin{figure*}[t] \centering     
\includegraphics[width=\columnwidth]{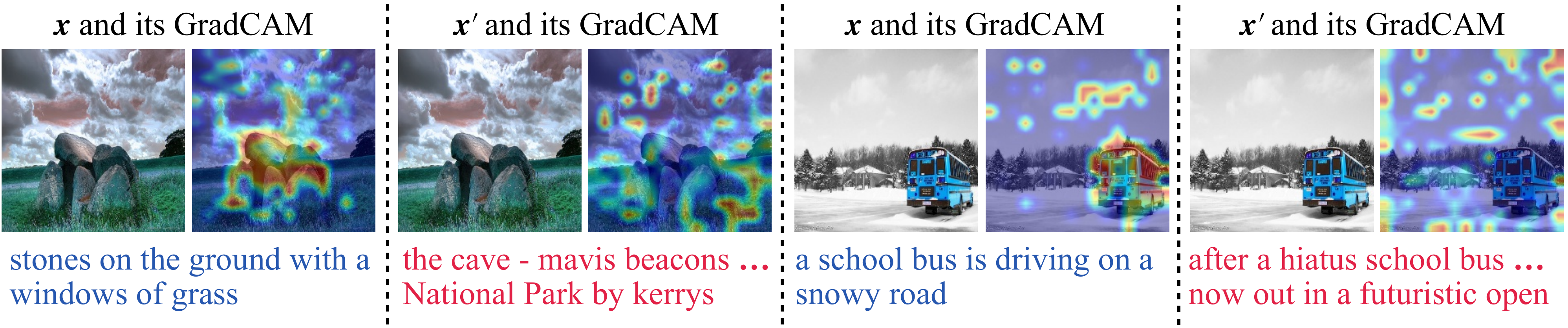}  
\vspace{-2em}
\caption{GradCAM for the original image $\bm{x}$ and our verbose counterpart $\bm{x}'$. The attention of our verbose images is more dispersed and uniform. We intercept only a part of the generated content. } 
\vspace{-1.5em}
\label{grad cam}
\end{figure*}

\textbf{Visual Interpretation.} 
We adopt GradCAM \citep{selvaraju2017grad}, a gradient-based visualization technique that generates attention maps highlighting the relevant regions in the input images for the generated sequences. From Fig. \ref{grad cam}, the attention of original images primarily concentrates on a local region containing a specific object mentioned in the generated caption. In contrast, our verbose images can effectively disperse attention and cause  VLMs to shift their focus from a specific object to the entire image region. Since the attention mechanism serves as a bridge between the input image and the output sequence of VLMs, we conjecture that the generation of a longer sequence can be reflected on an inaccurate focus and dispersed and uniform attention from the visual input.

\begin{table*}[t]
\caption{The $\text{CHAIR}_i$ (\%) and $\text{CHAIR}_s$ (\%) of the original images and our verbose images against four VLMs. Our verbose images can induce VLMs to generate more hallucinated objects.}
\label{tab:object hallucination}
\centering
\small
\setlength\tabcolsep{9.2pt}{
\begin{tabular}{@{}l|cccc|cccc@{}}
\toprule
\multirow{3}{*}{VLMs} & \multicolumn{4}{c|}{$\text{CHAIR}_i$ (\%)} & \multicolumn{4}{c}{$\text{CHAIR}_s$ (\%)} \\
& \multicolumn{2}{c}{MS-COCO} & \multicolumn{2}{c|}{ImageNet} & \multicolumn{2}{c}{MS-COCO} & \multicolumn{2}{c}{ImageNet}  \\ 
& Original & Ours  & Original & Ours & Original & Ours  & Original & Ours \\
\midrule 
BLIP & 11.41 & 79.93 & 22.29 & 89.80 & 12.77 & 84.22 & 13.77 & 90.33 \\
BLIP-2 & 12.03 & 52.30 & 25.30 & 69.83 & 10.99 & 35.02 & 11.77 & 46.11 \\
InstructBLIP & 23.66 & 55.56 & 40.11 & 69.27 & 38.04 & 75.46 & 34.55 & 64.55 \\
MiniGPT-4 & 19.42 & 46.65 & 29.20 & 65.50 & 19.61 & 52.01 & 16.57 & 54.37 \\
\bottomrule
\end{tabular}}
\end{table*}

\begin{table*}[t]
\begin{minipage}{\textwidth}
\caption{The length of generated sequences, energy consumption (J), and latency time (s) against BLIP-2 in different combinations of three loss objectives. }
\label{tab:ablation for loss functions}
\centering
\small
\setlength\tabcolsep{12.7pt}{
\begin{tabular}{@{}ccc|ccc|ccc@{}}
\toprule
\multirow{2}{*}{$\mathcal{L}_{1}$} & \multirow{2}{*}{$\mathcal{L}_{2}$} & \multicolumn{1}{c|}{\multirow{2}{*}{$\mathcal{L}_{3}$}} & \multicolumn{3}{c|}{MS-COCO} & \multicolumn{3}{c}{ImageNet} \\ 
&  & & Length & Latency & Energy & Length & Latency & Energy  \\
\midrule 
\checkmark &  &  & 119.46 &  3.96 & 162.40 & 147.87 & 4.52 & 185.64  \\
& \checkmark &  & 139.54 & 4.65 & 194.17 & 161.46 & 5.69 & 240.25  \\
&  & \checkmark & 104.03 &  3.29 & 135.75 & 129.02 & 3.90 & 161.87  \\
\checkmark & \checkmark &  & 177.95 & 6.47 & 267.01 & 217.78 & 7.47 & 306.09  \\
\checkmark &  & \checkmark & 150.79 & 4.51 & 182.16 & 151.57 & 4.71 & 194.40  \\
& \checkmark & \checkmark & 176.53 & 6.05 & 254.30 & 206.43 & 7.50 & 304.06  \\
\checkmark & \checkmark & \checkmark & \textbf{226.72} & \textbf{7.97} & \textbf{321.59} & \textbf{250.72} & \textbf{10.26} & \textbf{398.58} \\
\bottomrule
\end{tabular}}
\vspace{-0.5em}
\end{minipage}
\begin{minipage}{\textwidth}
\caption{The length of generated sequences, energy consumption (J), and latency time (s) against BLIP-2 in different combinations of two optimization modules. }
\label{tab:ablation for the optimization}
\centering
\small
\setlength\tabcolsep{9.65pt}{
\begin{tabular}{@{}cc|ccc|ccc@{}}
\toprule
\multirow{2}{*}{Temporal decay} & \multicolumn{1}{c|}{\multirow{2}{*}{Momentum}} & \multicolumn{3}{c|}{MS-COCO} & \multicolumn{3}{c}{ImageNet} \\  
& & Length & Latency & Energy & Length & Latency & Energy  \\
\midrule 
 &  &  152.49 & 4.70 & 205.09 & 144.90 & 5.31 & 231.83 \\
\checkmark &  & 199.92 & 7.02 & 292.55 & 231.03 & 7.88 & 318.34 \\
 & \checkmark & 187.32 & 6.89 & 274.67 & 214.92 & 7.49 & 308.11 \\
\checkmark & \checkmark  & \textbf{226.72} & \textbf{7.97} & \textbf{321.59} & \textbf{250.72} & \textbf{10.26} & \textbf{398.58} \\
\bottomrule
\end{tabular}}
\end{minipage}
\vspace{-0.5em}
\begin{minipage}{\textwidth}
\caption{The length of generated sequences, energy consumption (J), and latency time (s) against BLIP-2 with different perturbation magnitudes $\epsilon$.}
\label{tab:varying perturbation magnitude}
\centering
\small
\setlength\tabcolsep{9.2pt}{
\begin{tabular}{@{}c|cccc|cccc@{}}
\toprule
\multirow{2}{*}{Magnitude} & \multicolumn{4}{c|}{MS-COCO} & \multicolumn{4}{c}{ImageNet} \\ 
& Length & Latency & Energy & LIPIS & Length & Latency & Energy & LIPIS \\
\midrule 
2 & 91.75 & 3.06 & 126.22 & 0.0037 & 103.51 & 3.50 & 144.30 & 0.0038 \\
4 & 141.46 & 4.63 & 187.14 & 0.0121 & 147.30 & 4.90 & 199.24 & 0.0130 \\
8 & 226.72 & 7.97 & 321.59 & 0.0362 & 250.72 & 10.26 & 398.58 & 0.0372 \\
16 & 251.09 & 8.41 & 355.00 & 0.0879 & 272.95 & 9.51 & 380.86 & 0.0862 \\
32 & 287.22 & 9.13 & 377.64 & 0.1608 & 321.65 & 10.61 &  429.77 & 0.1575 \\
\bottomrule
\end{tabular}}
\end{minipage}
\end{table*}

\textbf{Textual Interpretation.} 
We investigate object hallucination in generated sequences using CHAIR \citep{rohrbach2018object}. 
$\text{CHAIR}_i$ is calculated as the fraction of hallucinated object instances, while $\text{CHAIR}_s$ represents the fraction of sentences containing a hallucinated object, with the results presented in Table \ref{tab:object hallucination}.
Compared to original images, which exhibit a lower object hallucination rate, the longer sequences produced by our verbose images contain a broader set of objects. This observation implies that our verbose images can prompt VLMs to generate sequences that include objects not present in the input image, thereby leading to longer sequences and higher energy-latency cost. \blue{Additionally, results of joint optimization of both images and texts, more results of visual interpretation, and additional discussions are provided in Appendix \ref{sec: joint optimization of both images and texts}, Appendix \ref{sec: visual interpretation}, and Appendix \ref{sec: Additional discussion}.}


\subsection{Ablation studies}
We explore the effect of the proposed three loss objectives, the effect of the temporal weight adjustment algorithm with momentum, and the effect of different perturbation magnitudes. 

\textbf{Effect of loss objectives.} Our verbose images consist of three loss objectives: $\mathcal{L}_{1}(\cdot)$, $\mathcal{L}_{2}(\cdot)$ and $\mathcal{L}_{3}(\cdot)$. To identify the individual contributions of each loss function and their combined effects on the overall performance, we evaluate various combinations of the proposed loss functions, as presented in Table \ref{tab:ablation for loss functions}. It can be observed that optimizing each loss function individually can generate longer sequences, and the combination of all three loss functions achieves the best results in terms of sequence length. This ablation study suggests that the three loss functions, which delay EOS occurrence, enhance output uncertainty, and improve token diversity, play a complementary role in extending the length of generated sequences.

\textbf{Effect of temporal weight adjustment.} During the optimization, we introduce two methods: a temporal decay for loss weighting and an addition of the momentum. As shown in Table \ref{tab:ablation for the optimization}, both methods contribute to the length of generated sequences. Furthermore, the longest length is obtained by combining temporal decay and momentum, providing a significant improvement over the baseline without both methods on MS-COCO and ImageNet datasets. It indicates that temporal decay and momentum can work synergistically to induce high energy-latency cost of VLMs.

\textbf{Effect of different perturbation magnitudes.} In our default setting, the perturbation magnitude $\epsilon$ is set as 8. To investigate the impact of different magnitudes, we vary $\epsilon$ under $[2,4,8,16,32]$ in Table \ref{tab:varying perturbation magnitude} and calculate the corresponding LIPIS \citep{zhang2018unreasonable} between  original images and their counterpart verbose images, which quantifies the perceptual difference. It can be observed that a larger perturbation magnitude $\epsilon$ results in a longer generated sequence by VLMs but produces more perceptible verbose images. Consequently, this trade-off between image quality and energy-latency cost highlights the importance of choosing an appropriate perturbation magnitude during evaluation. Additional ablation studies are shown in Appendix \ref{sec: Additional ablation study} and  
in Appendix \ref{sec: Grid search}.



\section{Conclusion}
In this paper, we aim to craft an imperceptible perturbation to induce high energy-latency cost of VLMs during the inference stage. We propose verbose images to prompt VLMs to generate as many tokens as possible. To this end, a delayed EOS loss, an uncertainty loss, a token diversity loss,  and a temporal weight adjustment algorithm are proposed to generate verbose images. Extensive experimental results demonstrate that, compared to original images, our verbose images can increase the length of generated sequences by 7.87$\times$ and 8.56$\times$ on MS-COCO and ImageNet across four VLMs. We hope that our verbose images can serve as a baseline for inducing high energy-latency cost of VLMs. Additional examples of our verbose images are shown in Appendix \ref{sec: Visualization}.

%


\subsection*{ACKNOWLEDGEMENT}
This work is supported in part by the National Natural Science Foundation of China under Grant 62171248, Shenzhen Science and Technology Program (JCYJ20220818101012025), and the PCNL KEY project (PCL2023AS6-1). This work is also supported by the UKRI grant: Turing AI Fellowship EP/W002981/1, EPSRC/MURI grant: EP/N019474/1. We would also like to thank the Royal Academy of Engineering and FiveAI.

\subsection*{ETHICS STATEMENT}
Please note that we restrict all experiments in the laboratory environment and do not support our verbose images in the real scenario.
The purpose of our work is to raise the awareness of the security concern in availability of VLMs and call for  practitioners to pay more attention to the energy-latency cost of VLMs and model trustworthy deployment.


\bibliography{iclr2024_conference}
\bibliographystyle{iclr2024_conference}

\clearpage
\appendix
{\bfseries\LARGE Appendix}
\vspace{1em}

\textbf{Overview.} The implementation details are described in Appendix \ref{sec: Implementation details}, including the introduction of target models in Appendix \ref{sec: Target models} and the experimental setups in Appendix \ref{sec: Experimental setups}. The transferability of our verbose images in the black-box setting is studied in Appendix \ref{sec: black box}, assuming that the victim VLMs are inaccessible. In addition to the captioning task, we conduct further experiments on other multi-modal tasks, including visual question answering (VQA) and visual reasoning, in Appendix \ref{sec: More tasks}. More results of length distribution are shown in Appendix \ref{sec: More results of length distribution}. The results of the joint optimization of both images and texts are demonstrated in Appendix \ref{sec: joint optimization of both images and texts}. Additional discussions on visual interpretation are provided in Appendix \ref{sec: visual interpretation}. \blue{Discussions on the feasibility analysis of an intuitive solution to study whether limitation on generation length can address the energy-latency vulnerability, the model performance of three energy-latency attacks, image embedding distance between original images and verbose counterpart, energy consumption for generating one verbose image, and standard deviation results of our main table are shown in Appendix \ref{sec: Additional discussion}.} Ablation studies on different sampling policies and maximum lengths of generated sequences are presented in Appendix \ref{sec: Additional ablation study}. The results of grid search for various parameters of loss weights and momentum values are reported in Appendix \ref{sec: Grid search}. Lastly, visual examples of original images and our verbose images against four VLM models are showcased in Appendix \ref{sec: Visualization}.

\section{Implementation details}
\label{sec: Implementation details}
In summary, we use the PyTorch framework \citep{paszke2019pytorch} and the LAVIS library \citep{li2023lavis} to implement the experiments. Note that every experiment is run on one NVIDIA Tesla A100
GPU with 40GB memory.

\subsection{Target models} 
\label{sec: Target models}
For ease of reproduction, we adopt four open-sourced VLMs as our target model and the implementation details of them are described as follows.

\textbf{Settings for BLIP.} We employ the BLIP with the basic multimodal mixture of an encoder-decoder model in 224M version. Following \citet{li2022blip}, we set the image resolution to 384 $\times$ 384, and a placeholder $\emptyset$ serves as the input text $\bm{c}_{\text{in}}$ of BLIP for the image captioning task.

\textbf{Settings for BLIP-2.} We utilize the BLIP-2 with an OPT-2.7B LM \citep{zhang2022opt}. As suggested in \citet{li2023blip}, the image resolution is 224 $\times$ 224, and a placeholder $\emptyset$ also serves as the input text $\bm{c}_{\text{in}}$ of BLIP-2 for the image captioning task.

\textbf{Settings for InstructBLIP.} We choose InstructBLIP with a Vicuna-7B LM \citep{chiang2023vicuna}. Following \citet{dai2023instructblip}, we set the image resolution to 224 $\times$ 224, and based on the instruction templates for the image captioning task provided in \citet{dai2023instructblip}, we configure the input text $\bm{c}_{\text{in}}$ of InstructBLIP accordingly as:
\textit{$<$Image$>$ What is the content of this image?}

\textbf{Settings for MiniGPT-4.} We adopt MiniGPT-4 with a Vicuna-7B LM \citep{chiang2023vicuna}. As suggested in \citet{zhu2023minigpt}, the image resolution is 224 $\times$ 224, and considering the predefined instruction templates for the image captioning task provided in \citet{zhu2023minigpt}, the input text $\bm{c}_{\text{in}}$ of MiniGPT-4 is set as: 

\textit{Give the following image: $<$Img$>$ImageContent$<$/Img$>$. You will be able to see the image once I provide it to you. Please answer my questions. $\#\#\#$Human: $<$Img$>$$<$ImageFeature$>$$<$/Img$>$ What is the content of this image? $\#\#\#$Assistant:}

\subsection{Experimental setups} 
\label{sec: Experimental setups}
\textbf{Setups for main experiments.} We perform the projected gradient descent (PGD) \citep{madry2017towards} algorithm to optimize sponge samples, NICGSlowDown, and our verbose images. Specifically, the optimization iteration is set as $T=1,000$, the perturbation magnitude is set as $\epsilon=8$ within $l_{\infty}$ restriction \citep{goodfellow2014explaining,carlini2019evaluating,xu2020adversarial,li2022semi,bai2020targeted,bai2021improving,bai2021targeted,bai2022hardly,bai2022improving,bai2022practical,gu2022vision,gu2022segpgd,liu2022watermark,wang2022triangle,wu2023defenses,he2023generating}, and the step size is set as $\alpha=1$. Besides, for the VLMs, we set the maximum length of generated sequences as $512$ and use nucleus sampling \citep{holtzman2019curious} with $p=0.9$ and temperature $t=1$ to sample the output sequences. For simplicity, we only consider a one-round conversation between the user and the VLMs. For the optimization of our verbose images, the parameters of loss weights is set as $a_1=10$, $b_1=-20$, $a_2=0$, $b_2=0$, $a_3=0.5$, and $b_3=1$ and the momentum is set as $0.9$.

\textbf{Setups for discussions.}
For the CHAIR \citep{rohrbach2018object}, it measures the extent of the object hallucination. A higher CHAIR value indicates the presence of more hallucinated objects in the sequence. As the calculation of CHAIR requires the object ground truth of an image, we employ the SEEM \citep{zou2023segment} method to segment each image and obtain the objects they contain.


For the results of the black-box setting described in Appendix \ref{sec: black box}, we leverage the transferability of the verbose images to induce high energy-latency cost. Specifically, we consider BLIP, BLIP-2, InstructBLIP, and MiniGPT-4 as the target victim VLMs, while the surrogate model is chosen as any VLM other than the target victim itself.

\begin{table*}[t]
\caption{The length of generated sequences, energy consumption (J), and latency time (s) of black-box  transferability across four VLMs of our verbose images. Our verbose images can transfer across different VLMs. }
\label{tab:transferability of four models of our verbose images}
\centering
\small
\setlength\tabcolsep{9.3pt}{
\begin{tabular}{@{}ll|ccc|ccc@{}}
\toprule
\multirow{2}{*}{Source model} & \multicolumn{1}{l|}{\multirow{2}{*}{Target model}} & \multicolumn{3}{c|}{MS-COCO} & \multicolumn{3}{c}{ImageNet} \\
 & & Length & Latency & Energy & Length & Latency & Energy  \\
\midrule 
None & \multirow{5}{*}{BLIP} & 10.03 &  0.21 & 9.51 & 10.17 & 0.22 & 9.10 \\
BLIP &  &  \textbf{318.66} & \textbf{5.13} & \textbf{406.65} & \textbf{268.25} & \textbf{4.31} & \textbf{344.91} \\
BLIP-2 &  & 14.51 & 0.24 & 10.05 & 14.03 & 0.24 & 10.23 \\
InstructBLIP &  &  63.43 & 2.84 & 142.46 & 54.14 & 2.52 & 131.22 \\
MiniGPT-4 &  & 48.50 & 10.23 & 316.28 & 49.14 & 10.20 & 321.29 \\
\midrule 

None & \multirow{5}{*}{BLIP-2} & 8.82 & 0.39 & 16.08 & 8.11 & 0.37 & 15.39 \\
BLIP &  & 36.09 & 1.19 &  47.07 & 73.22 & 2.39 & 99.24 \\
BLIP-2 &  & \textbf{226.72} & \textbf{7.97} & \textbf{321.59} & \textbf{250.72} & \textbf{10.26} & \textbf{398.58} \\
InstructBLIP &  & 140.05 & 3.91 & 166.40 & 145.39 & 4.07 & 175.01 \\
MiniGPT-4 &  & 140.88 & 3.81 & 154.43 & 140.92 & 3.91 & 165.95 \\
\midrule 

None & \multirow{5}{*}{InstructBLIP} & 63.79 & 2.97 & 151.80 & 54.40 & 2.60 & 128.03 \\
BLIP &   & 91.94 & 4.13 & 203.94 & 82.66 & 3.77 & 186.51 \\
BLIP-2 &  & 109.01 & 4.87 & 240.30 & 99.25 & 4.53 & 225.55 \\
InstructBLIP &  & \textbf{140.35} & \textbf{6.15} & \textbf{316.06} & \textbf{131.79}  & \textbf{6.05} & \textbf{300.43} \\
MiniGPT-4 &  & 100.08 & 4.42 & 210.58 & 99.42 & 4.47 & 219.08 \\
\midrule 

None & \multirow{5}{*}{MiniGPT-4} & 45.29 & 10.39 &  329.50 & 40.93 &  9.11 & 294.68 \\
BLIP &  & 229.10 & 48.90 & 1562.25 & 254.57 & 54.57 & 1691.51\\
BLIP-2 &  & 296.77 & 58.84 & 1821.66 & 289.19 & 58.79 & 1826.81 \\
InstructBLIP &  & 270.73 & 48.88 & 1551.04 & 258.32 & 50.26 & 1632.01 \\
MiniGPT-4 &   & \textbf{321.35} & \textbf{67.14} & \textbf{2113.29} & \textbf{321.24} & \textbf{64.31} &  \textbf{2024.62} \\
\bottomrule
\end{tabular}}
\vspace{-1em}
\end{table*}

\begin{table*}[t]
\caption{The length of generated sequences, energy consumption (J), and latency time (s) against BLIP-2 on VQA and visual reasoning. Our verbose images can still achieve better on VQA and visual reasoning.}
\label{tab:different tasks on vqa and reason}
\centering
\small
\setlength\tabcolsep{4pt}{
\begin{tabular}{@{}l|ccc|ccc|ccc@{}}
\toprule
\multirow{2}{*}{Attacking method} & \multicolumn{3}{c|}{Image Caption} & \multicolumn{3}{c|}{VQA} & \multicolumn{3}{c}{Visual Reason}\\
 & Length & Latency & Energy & Length & Latency & Energy & Length & Latency & Energy \\
\midrule 
Original & 8.82 & 0.39 & 16.08 & 6.43 & 0.44 & 17.92 & 5.70 & 0.44 & 18.23 \\
Noise & 9.55 & 0.43 & 17.53 & 6.62 & 0.45 & 17.09 & 6.70 & 0.53 & 19.83 \\
Sponge samples & 22.53 & 0.73 & 30.20 & 133.24 & 5.04 & 191.16 & 142.28 & 5.24 & 205.35 \\
NICGSlowDown & 103.54 & 3.78 & 156.61 & 127.96 & 5.07 & 190.13 & 136.44 & 5.21 & 185.88 \\
\textbf{Verbose images (Ours)} & \textbf{226.72} & \textbf{7.97} & \textbf{321.59} & \textbf{271.95} & \textbf{11.49} & \textbf{365.49} & \textbf{280.10} & \textbf{12.52} & \textbf{413.06}  \\
\bottomrule
\end{tabular}}
\end{table*}

\section{Black-box setting}
\label{sec: black box}
In the previous experiments, we assume that the victim VLMs are fully accessible. In this section, we consider a more realistic scenario, where the victim VLMs are unknown \citep{ilyas2018black,bai2020improving}. 
To induce high energy-latency cost of black-box VLMs, we can leverage the transferability property \citep{dong2018boosting} of our verbose images. We can first craft verbose images on a known and accessible surrogate model and then utilize them to transfer to the target victim VLM. The black-box transferability  results across four VLMs of our verbose images are evaluated in Table \ref{tab:transferability of four models of our verbose images}. When the source model is set as `None', it indicates that we evaluate the energy-latency cost for  the target model using the original images. The results show that our transferable verbose images are less effective than the white-box verbose images but still result in a longer generated sequence. 

\begin{figure*}[t] \centering    
\begin{minipage}{\textwidth}
\subfigure[BLIP] { 
\label{BLIP of length distribution}  
\includegraphics[width=0.232\columnwidth]{PDFs/main_results/blip_coco.pdf}  
}    
\subfigure[BLIP-2] {
\label{BLIP2 of length distribution}  \includegraphics[width=0.232\columnwidth]{PDFs/main_results/opt_coco.pdf}  
}     
\subfigure[InstructBLIP] { 
\label{InstructBLIP of length distribution}  \includegraphics[width=0.232\columnwidth]{PDFs/main_results/instr_coco.pdf}  
}    
\subfigure[MiniGPT-4] { 
\label{MiniGPT-4 of length distribution}  \includegraphics[width=0.232\columnwidth]{PDFs/main_results/gpt_coco.pdf}  
}    
\caption{The length distribution of four VLM models on MS-COCO dataset, including (a) BLIP. (b) BLIP-2. (c) InstructBLIP. (d) MiniGPT-4. The peak of length distribution of our verbose images shift towards longer sequences.} 
\label{all length distribution of coco}
\end{minipage}
\begin{minipage}{\textwidth}
\subfigure[BLIP] { 
\label{BLIP of length distribution}  
\includegraphics[width=0.232\columnwidth]{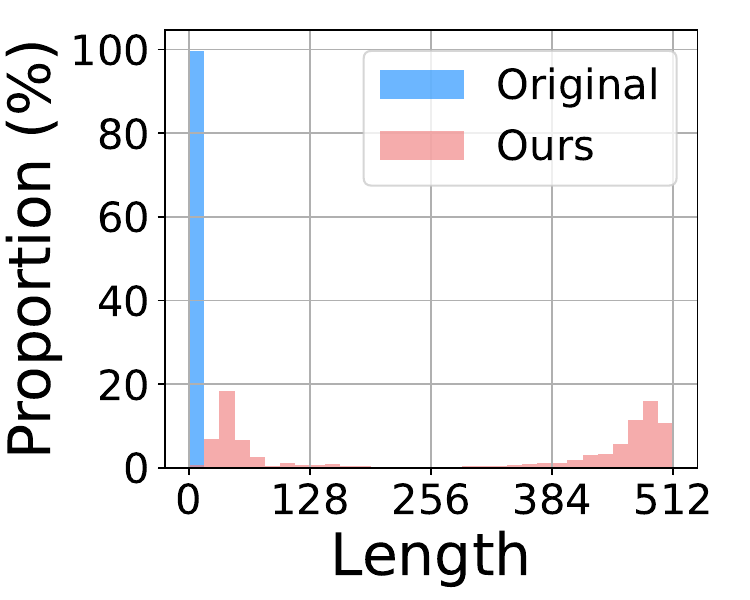}  
}    
\subfigure[BLIP-2] {
\label{BLIP2 of length distribution}  \includegraphics[width=0.232\columnwidth]{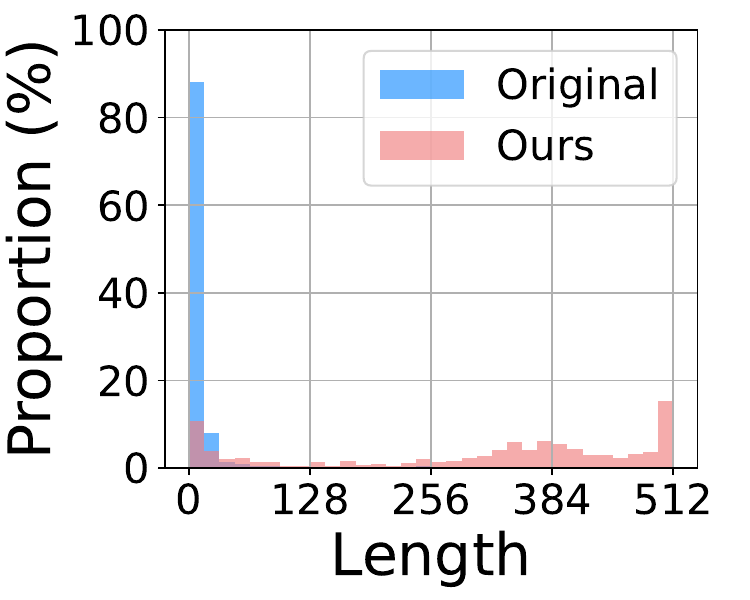}  
}     
\subfigure[InstructBLIP] { 
\label{InstructBLIP of length distribution}  \includegraphics[width=0.232\columnwidth]{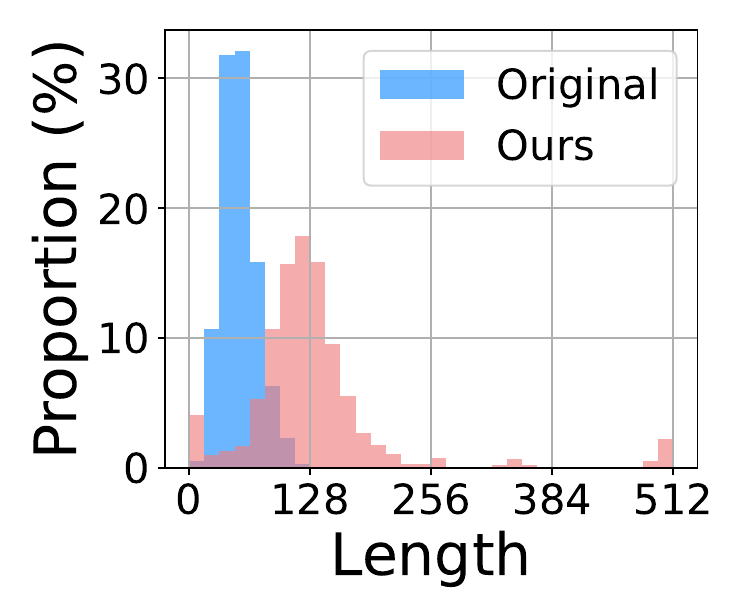}  
}    
\subfigure[MiniGPT-4] { 
\label{MiniGPT-4 of length distribution}  \includegraphics[width=0.232\columnwidth]{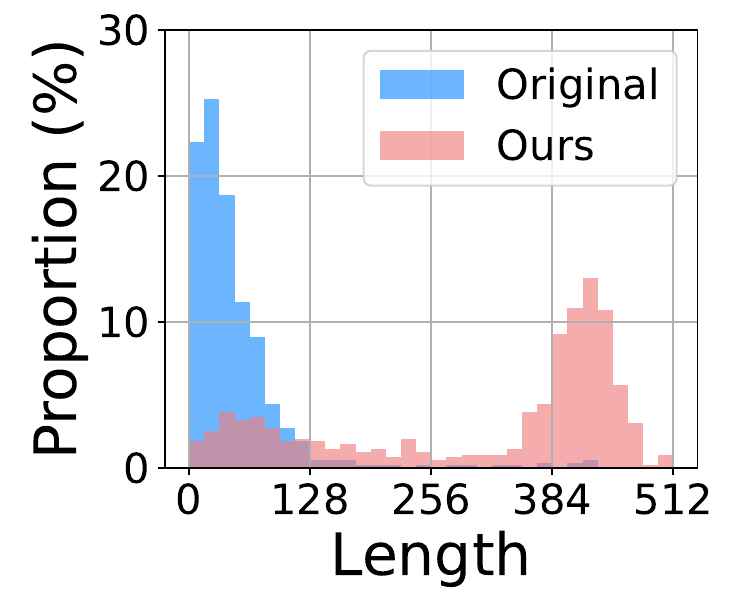}  
}    
\caption{The length distribution of four VLM models on ImageNet dataset, including (a) BLIP. (b) BLIP-2. (c) InstructBLIP. (d) MiniGPT-4. The peak of length distribution of our verbose images shift towards longer sequences.} 

\label{all length distribution of imagenet}
\end{minipage}
\end{figure*}

\section{More tasks}
\label{sec: More tasks}
To verify the effectiveness of our verbose images, we induce high energy-latency cost on two additional multi-modal tasks: visual question answering (VQA) and visual reasoning. Following \citet{li2023blip}, we use VQAv2 dataset \citep{goyal2017making} for VQA and GQA dataset \citep{hudson2019gqa} for visual reasoning. We use BLIP-2 as the target model, and following the recommendations in \citet{li2023blip}, we set the prompt template as ``\textit{Question: \{\} Answer:}". Unless otherwise specified, other settings remain unchanged. Table \ref{tab:different tasks on vqa and reason} demonstrates that our verbose images can induce the highest energy-latency cost among three multi-modal tasks.

\section{More results of length distribution}
\label{sec: More results of length distribution}
We provide more results of the length distribution on MS-COCO dataset in Fig. \ref{all length distribution of coco} and ImageNet dataset in Fig. \ref{all length distribution of imagenet}. \blue{The length distribution of generated sequences in the four VLM models exhibits a bimodal distribution. Specifically, our verbose images tend to prompt VLMs to generate either long or short sequences. We conjecture the reasons as follows. A majority of the long sequences are generated, confirming the effectiveness of our verbose images. As for the short sequences, we have carefully examined the generated content and observed two main cases. In the first case, our verbose images fail to induce long sentences, particularly for BLIP and BLIP-2, which lack instruction tuning and have smaller parameters. In the second case, our verbose images can confuse the VLMs, leading them to generate statements such as `I am sorry, but I cannot describe the image.' This scenario predominantly occurs with InstructBlip and MiniGPT-4, both of which have instruction tuning and larger parameters.
}

\section{joint optimization of both images and texts}
\label{sec: joint optimization of both images and texts}
VLMs combine vision transformers and large language models to obtain an enhanced zero-shot performance in multi-modal tasks \citep{liu2023visual,li2021align,li2023blip,ma2022visual,ma2023follow}. Hence, VLMs are capable of processing both visual and textual inputs, enabling them to handle multi-modal tasks effectively. In this section, we will adopt our proposed losses and the temporal weight adjustment algorithm to optimize both the  imperceptible perturbation of visual inputs and tokens of textual inputs to induce high energy-latency cost of VLMs. For the optimization of textual inputs, we update a parameterized distribution matrix to optimize input textual tokens, as suggested in \citet{guo2021gradient}. The number of optimized tokens is set as 8. Besides, Adam optimizer \citep{kingma2014adam} with a learning rate of 0.5 is used to optimize input textual tokens every iteration. Moreover, both the  imperceptible perturbation of visual inputs and tokens of textual inputs are jointly optimized.
Unless otherwise specified, other settings remain unchanged. Table \ref{tab:joint optimization of both images and texts} demonstrates that our methods can still induce VLMs to generate longer sequences than other methods under the joint optimization of both images and texts of VLMs.


\section{Visual interpretation}
\label{sec: visual interpretation}
\begin{figure*}[t] \centering   
\begin{minipage}{\textwidth}
    \includegraphics[width=\columnwidth]{PDFs/grad_cam.pdf}  
\end{minipage}
\begin{minipage}{\textwidth}
    \includegraphics[width=\columnwidth]{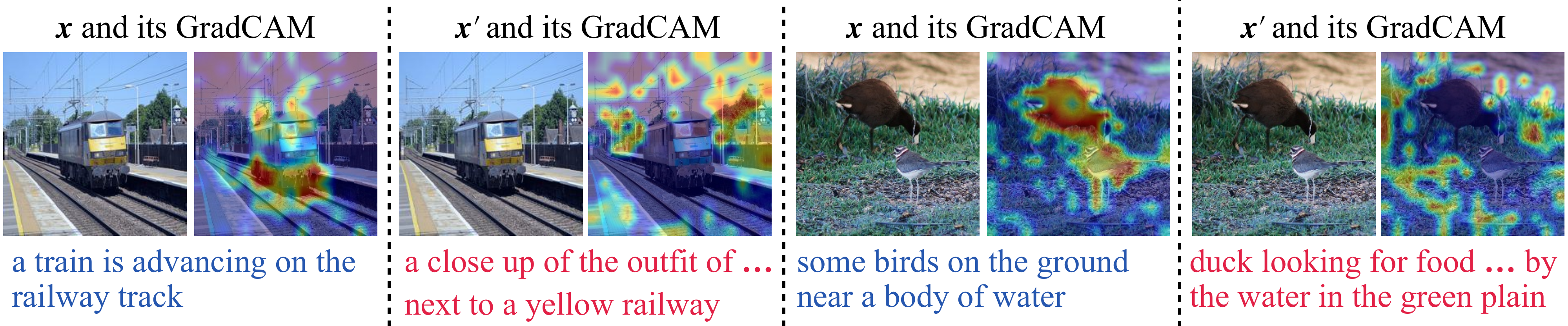}  
\end{minipage}
\begin{minipage}{\textwidth}
    \includegraphics[width=\columnwidth]{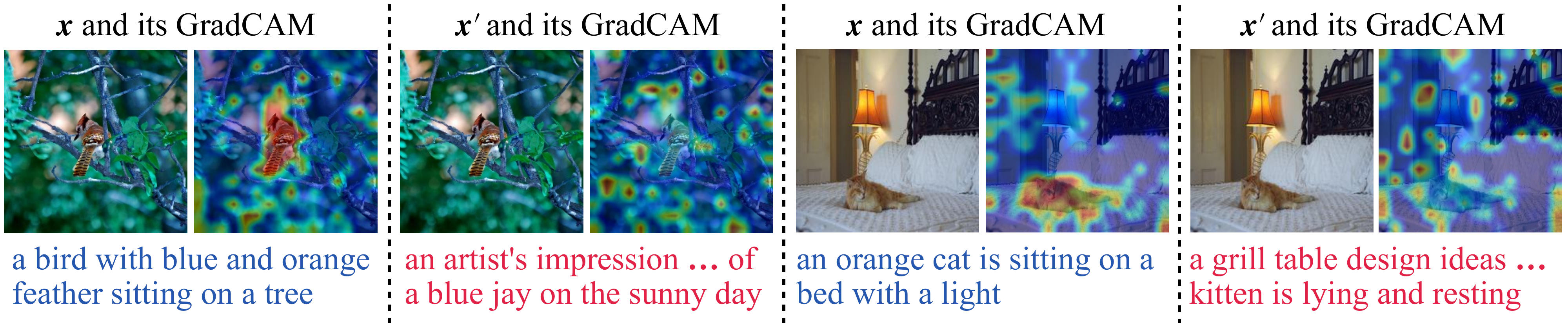}  
\end{minipage}
\caption{GradCAM for the original image $\bm{x}$ and our verbose counterpart $\bm{x}'$. The attention of our verbose images is more dispersed and uniform. Note that we intercept only a part of the generated content. } 
\label{all grad cam}
\end{figure*}

We show more visual interpretation results of the original images and our verbose images by using Grad-CAM \citep{selvaraju2017grad}. The results are demonstrated in Fig. \ref{all grad cam}.

\begin{table*}[t]
\caption{The length of generated sequences, energy consumption (J), and latency time (s) against BLIP-2 of our verbose images by the joint optimization of both images and texts. Our verbose images can still achieve better.}
\label{tab:joint optimization of both images and texts}
\centering
\small
\setlength\tabcolsep{12.3pt}{
\begin{tabular}{@{}l|ccc|ccc@{}}
\toprule
\multirow{2}{*}{Attacking method} & \multicolumn{3}{c|}{MS-COCO} & \multicolumn{3}{c}{ImageNet} \\
 & Length & Latency & Energy & Length & Latency & Energy  \\
\midrule 
Original & 5.81 & 0.37 & 14.81 & 5.85 & 0.35 & 13.48 \\
Noise & 5.91 & 0.38 & 15.02 & 6.65 & 0.39 & 16.18 \\
Sponge samples & 162.78 & 4.66 & 193.76 & 190.52 & 5.53 & 223.94 \\
NICGSlowDown & 185.21 & 5.66 & 227.03 & 200.51 & 6.71 & 286.09 \\
\textbf{Verbose images (Ours)} & \textbf{270.25} & \textbf{9.39} & \textbf{378.14} &  \textbf{262.92} &  \textbf{9.03} & \textbf{365.28} \\
\bottomrule
\end{tabular}}
\end{table*}

\begin{figure*}[t] \centering   
\begin{minipage}{\textwidth}
    \includegraphics[width=\columnwidth]{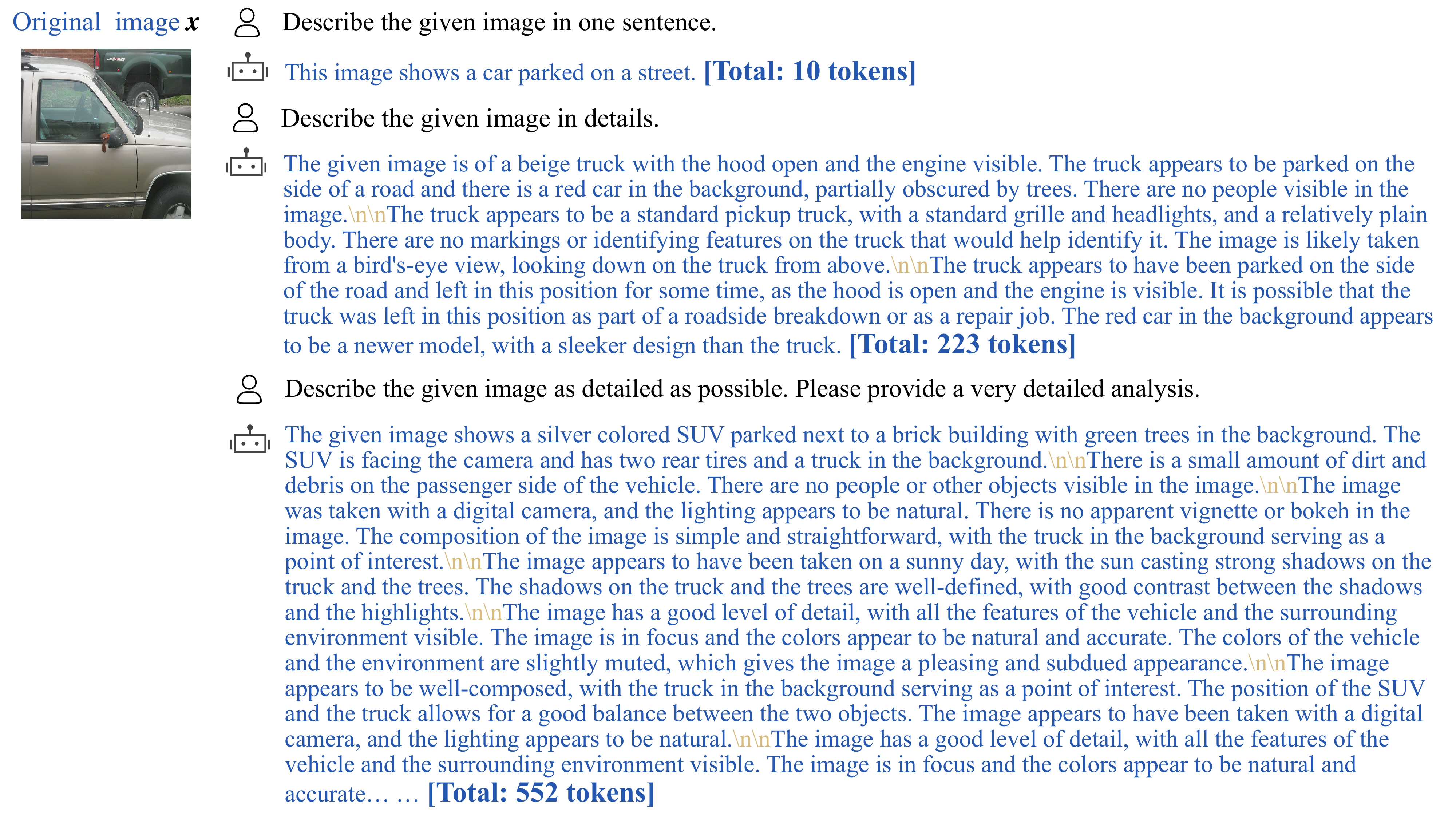}  
\end{minipage}
\caption{\blue{An example of generated sequences from MiniGPT-4 by different input prompts. Users have diverse requirements and input data, leading to a wide range of lengths of generated sequences.}} 
\label{different text prompts introduce different generation length}
\end{figure*}

\section{\blue{Additional discussions}}
\label{sec: Additional discussion}
\blue{We conduct additional discussions, including the feasibility analysis of an intuitive solution to study whether limitation on generation length can address the energy-latency vulnerability, the model performance of three energy-latency attacks, image embedding distance between original images and verbose counterpart, energy consumption for generating one verbose image, and standard deviation results of our main table.}

\subsection{\blue{feasibility analysis of an intuitive solution}}

\blue{An intuitive solution to mitigate the energy-latency vulnerability is to impose a limitation on generation length. We argue that such an intuitive solution is infeasible and the reason is as follows.}  

\blue{(1) Users have diverse requirements and input data, leading to a wide range of sentence lengths and complexities. For example, the prompt text of `Describe the given image in one sentence.' and `Describe the given image in details.' can introduce different lengths of generated sentences. We visualize a case in Fig. \ref{different text prompts introduce different generation length}. Consequently, service providers often consider a large token limit to accommodate these diverse requirements and ensure that the generated sentences are complete and meet users' expectations. Previous work, NICGSlowDown \citep{chen2022nicgslowdown}, also states the same view as us. Besides, as shown in Table \ref{tab:different tasks on vqa and reason}, the results can demonstrate that our verbose images are adaptable for different prompt texts and can induce the length of generated sentences closer to the token limit set by the service provider. As a result, the energy-latency cost can be increased while staying within the imposed constraints.  }

\blue{(2) We argue that this attack surface about availability of VLMs becomes more important and our verbose images can induce more serious attack consequences in the era of large (vision) language models. The development of VLMs and LLMs has led to models capable of generating longer sentences with logic and coherence. Consequently, service providers have been increasing the maximum allowed length of generated sequences to ensure high-quality user experiences. For instance, gpt-3.5-turbo and gpt-4-turbo allow up to 4,096 and 8,192 tokens, respectively. Hence, we would like to uncover that while longer generated sequences can indeed improve service quality, they also introduce potential security risks about energy-latency cost, as our verbose images demonstrates. Therefore, when VLM service providers consider increasing the maximum length of generated sequences for better user experience, they should not only focus on the ability of VLMs but also take the maximum energy consumption payload into account.  }

\begin{table*}[t]
\begin{minipage}{\textwidth}
\caption{\blue{The attacking performance, including the length of generated sequences, energy consumption (J), and latency time (s), and captioning performance, including BLEU-1, BLEU-2, BLEU-3, BLEU-4, and CIDEr of sponge samples, NICGSlowDown, and our verbose images.}}
\label{tab:performance of BLUE and CIDEr}
\centering
\small
\setlength\tabcolsep{4.9pt}{
\begin{tabular}{@{}l|ccc|ccccc@{}}
\toprule
\multirow{2}{*}{Attacking method} & \multicolumn{3}{c|}{Attacking performance} & \multicolumn{5}{c}{Captioning performance} \\ 
& Length & Latency & Energy & BLEU-1 & BLEU-2 & BLEU-3 & BLEU-4 & CIDEr \\
\midrule 
Sponge samples & 22.53 & 0.73 & 30.20 & 0.298 & 0.193 & 0.120 & 0.073 & 0.571 \\
NICGSlowDown & 103.54 & 3.78 & 156.61 & 0.012 & 0.005 & 0.002 & 0.001 & 0.027 \\
\textbf{Verbose images (Ours)} & 226.72 & 7.97 & 321.59 & 0.026 & 0.012 & 0.005 & 0.003 & 0.099  \\
\bottomrule
\end{tabular}}
\end{minipage}
\end{table*}

\begin{table*}[t]
\begin{minipage}{\textwidth}
\caption{\blue{The image embedding distance between original images and attacked counterpart of sponge samples, NICGSlowDown, and our verbose images. All these methods are similar in image embedding distance. }}
\label{tab:image embedding distance}
\centering
\small
\setlength\tabcolsep{2.3pt}{
\begin{tabular}{@{}l|cccc|cccc@{}}
\toprule
\multirow{2}{*}{Attacking method} & \multicolumn{4}{c|}{MS-COCO} & \multicolumn{4}{c}{ImageNet} \\ 
& BLIP & BLIP-2 & InstructBLIP & MiniGPT-4 & BLIP & BLIP-2 & InstructBLIP & MiniGPT-4 \\
\midrule 
Sponge samples & 0.965 & 0.977 & 0.978 & 0.978 & 0.968 & 0.979 & 0.981 & 0.981
 \\
NICGSlowDown & 0.966 & 0.978 & 0.979 & 0.971 & 0.968 & 0.980 & 0.981 & 0.975
 \\
\textbf{Verbose images (Ours)} & 0.965 & 0.970 & 0.971 & 0.969 & 0.966 & 0.969 & 0.970 & 0.971
 \\
\bottomrule
\end{tabular}}
\end{minipage}
\end{table*}

\begin{table*}[t]
\begin{minipage}{\textwidth}
\caption{\blue{The length of generated sequences, energy consumption (J) and latency time (s) during attack, and energy consumption (J) and latency time (s) during generation against
BLIP-2 for generating one verbose image. The results indicate a positive correlation between the energy-latency cost during attack, the energy-latency cost during generation, and the number of attacking iterations.  }}
\label{tab:energy consumption for generating one attacked image}
\centering
\small
\setlength\tabcolsep{18pt}{
\begin{tabular}{@{}c|ccc|cc@{}}
\toprule
\multirow{2}{*}{Attacking iterations} & \multicolumn{3}{c|}{During attack} & \multicolumn{2}{c}{During generation} \\ 
& Length & Latency & Energy & Latency & Energy \\
\midrule 
0 & 8.40 & 0.45 & 17.91 & 0 & 0 \\
100 & 70.26 & 3.77 & 136.73 & 50.28	& 6593.69
 \\
500 & 175.24 & 6.84 & 285.63 & 261.20 & 36537.34
 \\
1000 & 226.72 & 7.97 & 321.59 & 442.12 & 67230.84
 \\
\bottomrule
\end{tabular}}
\end{minipage}
\end{table*}

\begin{table*}[t]
\caption{\blue{The standard deviation results for length of generated sequences, energy consumption (J), and latency time (s). }}
\label{tab:standard deviation results of our main table}
\centering
\small
\setlength\tabcolsep{6.75pt}{
\begin{tabular}{@{}ll|ccc|ccc@{}}
\toprule
\multirow{2}{*}{VLM model} & \multicolumn{1}{l|}{\multirow{2}{*}{Method}} & \multicolumn{3}{c|}{MS-COCO} & \multicolumn{3}{c}{ImageNet} \\
 & & Length & Latency & Energy & Length & Latency & Energy  \\
\midrule 
\multirow{5}{*}{BLIP} & Original & 1.96 & 0.16	& 6.97 & 1.83 & 0.17 & 7.13 \\
& Noise & 1.83 & 0.14 & 6.22 & 1.86 & 0.14 & 6.03  \\
& Sponge samples & 100.69 & 1.66 & 145.27 & 124.64 & 2.02 & 171.51  \\
& NICGSlowDown & 205.71 & 3.21 & 278.81 & 214.22 & 3.30 & 294.34  \\
& \textbf{Verbose images (Ours)} &  207.88 & 3.26 & 284.50 & 209.32 & 3.31 & 297.79 \\
\midrule
\multirow{5}{*}{BLIP-2} & Original & 3.25 & 0.20 & 8.67 & 3.32 & 0.22 & 8.97   \\
& Noise &  3.26 & 0.22 & 8.68 & 3.08 & 0.21 & 8.93  \\
& Sponge samples &  64.11 & 1.09 & 44.22 & 100.63 & 1.80 & 77.80  \\
& NICGSlowDown &  166.75 & 2.48 & 104.35 & 189.63 & 2.14 & 92.32  \\
& \textbf{Verbose images (Ours)} &  170.70	& 3.74 & 161.27 & 164.81 & 4.76 & 191.38  \\
\midrule
\multirow{5}{*}{InstructBLIP} & Original & 19.90 & 0.71 & 37.07 & 17.76 & 0.72 & 37.47  \\
& Noise &  18.79 & 0.68 & 33.93 & 18.31 & 0.71	& 35.02   \\
& Sponge samples &  23.84 & 0.70 & 37.21 & 23.36 & 0.63 & 32.35  \\
& NICGSlowDown &  22.86 & 0.70 & 36.01 & 24.29 & 0.67 & 33.43  \\
& \textbf{Verbose images (Ours)} & 56.52 & 1.07 & 56.65 & 80.55 & 1.62 & 85.84   \\
\midrule
\multirow{5}{*}{MiniGPT-4} & Original & 47.54 & 7.21 & 227.09 & 52.46 & 7.29 & 231.74  \\
& Noise &  47.93 & 6.79 & 212.87 & 53.30 & 6.83	& 219.14  \\
& Sponge samples &  239.17 & 10.02 & 330.83 & 196.57 & 11.04 & 353.66  \\
& NICGSlowDown &  264.48 & 11.36 & 382.07 & 188.33 & 11.98 & 379.11  \\
& \textbf{Verbose images (Ours)} &  223.39 & 11.53 & 352.66 & 160.35 & 11.10 & 352.03  \\
\bottomrule
\end{tabular}}
\end{table*}

\begin{table*}[t]
\caption{The length of generated sequences, energy consumption (J), and latency time (s) against BLIP-2 in different sampling policies. Our verbose images can still achieve better in different sampling policies. }
\label{tab:different sampling policies}
\centering
\small
\setlength\tabcolsep{6.1pt}{
\begin{tabular}{@{}ll|ccc|ccc@{}}
\toprule
\multirow{2}{*}{Sampling method} & \multicolumn{1}{l|}{\multirow{2}{*}{Method}} & \multicolumn{3}{c|}{MS-COCO} & \multicolumn{3}{c}{ImageNet} \\
 & & Length & Latency & Energy & Length & Latency & Energy  \\
\midrule 
\multirow{5}{*}{Greedy search} & Original & 8.19 & 0.27 & 10.06 & 7.58 & 0.24 & 10.23 \\
& Noise & 7.32 & 0.23 & 10.24 & 7.74 & 0.25 & 10.43 \\
& Sponge samples & 8.47 & 0.27 & 11.30 & 8.11 & 0.27 & 10.90 \\
& NICGSlowDown &  195.44 & 5.22 & 216.48 & 267.02 & 7.45 & 301.47 \\
& \textbf{Verbose images (Ours)} & \textbf{323.97} & \textbf{8.86} & \textbf{367.72} & \textbf{352.16} & \textbf{9.56} & \textbf{399.19}  \\
\midrule
\multirow{5}{*}{Beam search} & Original & 9.53 & 0.42 & 17.74 & 8.84 & 0.43 & 18.75 \\
& Noise & 9.34 & 0.41 & 16.87 & 8.76 & 0.45 & 19.36 \\
& Sponge samples & 9.96 & 0.46 & 18.36 & 8.96 & 0.39 & 17.99 \\
& NICGSlowDown & 305.73 & 8.88  & 451.59 & 425.87 & 12.20 & 610.64 \\
& \textbf{Verbose images (Ours)} & \textbf{437.02} & \textbf{12.40} & \textbf{639.74} & \textbf{469.87} & \textbf{13.00} & \textbf{683.11} \\
\midrule
\multirow{5}{*}{Top-k sampling} & Original & 8.20 & 0.33 & 13.22 & 7.54 & 0.31 & 12.85 \\
& Noise & 8.15 & 0.32 & 13.17 & 7.59 & 0.32 & 13.01 \\
& Sponge samples & 8.53 & 0.35 & 14.86 & 7.73 & 0.26 & 11.03 \\
& NICGSlowDown & 207.96 & 5.77 & 225.89 & 262.11 & 7.12 & 291.76 \\
& \textbf{Verbose images (Ours)} & \textbf{293.48} & \textbf{8.08} & \textbf{330.39} & \textbf{372.53} & \textbf{10.24} & \textbf{417.89} \\
\midrule
\multirow{5}{*}{Nucleus sampling} & Original & 8.82 & 0.39 & 16.08 & 8.11 & 0.37 & 15.39 \\
& Noise & 9.55 & 0.43 & 17.53 & 8.37 & 0.44 & 19.39 \\
& Sponge samples & 22.53 & 0.73 & 30.20 & 43.59 & 1.51 & 63.27 \\
& NICGSlowDown & 103.54 & 3.78 & 156.61 & 129.68 & 4.34 & 180.06 \\
& \textbf{Verbose images (Ours)} & \textbf{226.72} & \textbf{7.97} & \textbf{321.59} & \textbf{250.72} & \textbf{10.26} & \textbf{398.58} \\
\bottomrule
\end{tabular}}
\end{table*}

\begin{table*}[t]
\caption{The length of generated sequences, energy consumption (J), and latency time (s) against BLIP-2 in different maximum lengths. Our verbose images can still achieve better in different maximum lengths. }
\label{tab:different maximum length}
\centering
\small
\setlength\tabcolsep{6.2pt}{
\begin{tabular}{@{}ll|ccc|ccc@{}}
\toprule
\multirow{2}{*}{Maximum length} & \multicolumn{1}{l|}{\multirow{2}{*}{Method}} & \multicolumn{3}{c|}{MS-COCO} & \multicolumn{3}{c}{ImageNet} \\
 & & Length & Latency & Energy & Length & Latency & Energy  \\
\midrule 
\multirow{5}{*}{128} & Original & 7.88 & 0.31 & 13.34 & 8.61 & 0.35 & 14.49 \\
& Noise & 8.45 & 0.35 & 14.11 & 8.25 & 0.34 & 13.81\\
& Sponge samples & 16.60 &  0.51 & 21.79 & 22.73 & 0.72 & 28.90 \\
& NICGSlowDown & 35.94 & 1.26 & 49.60 & 40.46 & 1.46 & 59.13 \\
& \textbf{Verbose images (Ours)} & \textbf{61.55} & \textbf{2.25} & \textbf{90.53} &\textbf{66.89} & \textbf{2.38} & \textbf{91.97}
\\
\midrule
\multirow{5}{*}{256} & Original & 8.18 & 0.32 & 13.01 & 8.03 & 0.34 & 13.83  \\
& Noise & 8.26 & 0.35 & 14.67 & 8.02 & 0.34 & 13.46 \\
& Sponge samples & 18.75 & 0.55 & 23.42 &  31.68 & 0.93 & 37.26 \\
& NICGSlowDown & 60.73 & 2.00 & 80.83 & 70.57 & 2.21 & 87.41 \\
& \textbf{Verbose images (Ours)} & \textbf{109.13} & \textbf{4.04} & \textbf{163.74} & \textbf{116.94} & \textbf{4.17} & \textbf{167.02}  \\
\midrule
\multirow{5}{*}{512} & Original & 8.82 & 0.39 & 16.08 & 8.11 & 0.37 & 15.39 \\
& Noise & 9.55 & 0.43 & 17.53 & 8.37 & 0.44 & 19.39 \\
& Sponge samples & 22.53 & 0.73 & 30.20 & 43.59 & 1.51 & 63.27 \\
& NICGSlowDown & 103.54 & 3.78 & 156.61 & 129.68 & 4.34 & 180.06 \\
& \textbf{Verbose images (Ours)} & \textbf{226.72} & \textbf{7.97} & \textbf{321.59} & \textbf{250.72} & \textbf{10.26} & \textbf{398.58} \\
\midrule
\multirow{5}{*}{1024} & Original & 8.18 & 0.35 & 13.97 &  8.49 &  0.35 & 14.69 \\
& Noise & 7.78 & 0.40 & 14.94 & 8.10 & 0.33 & 14.39 \\
& Sponge samples & 30.60 & 0.92 & 34.34 & 64.10 & 1.97 &  85.01 \\
& NICGSlowDown & 193.97 & 5.80 & 237.39 & 256.35 & 7.71 & 312.95 \\
& \textbf{Verbose images (Ours)} & \textbf{417.66} & \textbf{13.92} & \textbf{586.40} & \textbf{447.88} & \textbf{14.81} & \textbf{611.60} \\
\bottomrule
\end{tabular}}
\end{table*}

\subsection{\blue{evaluation performance of energy-latency manipulation}}
\blue{We evaluate the BLEU and CIDEr scores for sponge samples, NICGSlowDown, and our verbose images on MS-COCO for BLIP-2, as illustrated in Table \ref{tab:performance of BLUE and CIDEr}. Concretely, our verbose images generate the longest sequence, extending it to 226.72, while sponge samples, despite their superior captioning performance, only increase the length to 22.53. Furthermore, both the length of the generated sequence and the captioning performance of our verbose images outperform those of NICGSlowDown. }

\subsection{\blue{image embedding distance between original images and attacked counterpart}}
\blue{We adopt the image encoder of CLIP to extract the image embedding. Then the image embedding distance is calculated as the cosine similarity of both original images and the corresponding sponge samples \citep{shumailov2021sponge}, NICGSlowDown \citep{chen2022nicgslowdown}, and our verbose images. The results are shown in Table \ref{tab:image embedding distance} which demonstrates that these methods are similar in image embedding distance.}

\subsection{\blue{energy consumption for generating one attacked image}}

\blue{We calculate the energy-latency cost for the generation of one verbose image and show the results in Table \ref{tab:energy consumption for generating one attacked image}. It can be observed that the energy-latency cost during attack increases with the number of attack iterations, along with an increase in energy consumption for generating one verbose image. This finding provides valuable insights into the positive relation between the attack performance and the energy consumption associated with the generation of verbose images. }

Besides, the overall energy-latency cost of generating a single verbose image is higher than that of using it to attack VLMs. Therefore, it is necessary for the attacker to make full use of every generated verbose image and learn from DDoS attack strategies to perform this attack more effectively. Specifically, the attacker can instantly send as many copies of the same verbose image as possible to VLMs, which increases the probability of exhausting the computational resources and reducing the availability of VLMs service. Once the attack is successful and causes the competitor's service to collapse, the attacker will acquire numerous users from the competitor and gain significant benefits, revealing the necessity of the application of deep learning in security-sensitive scenarios \citep{li2016mutual,li2024nearest,liu2006spatio,tang2004video,gong2013multi,zhang2019joint}.

\subsection{\blue{standard deviation results}}
\blue{The standard deviation results for length of generated sequences, energy consumption, and latency time are shown in Table \ref{tab:standard deviation results of our main table}.}

\section{Additional ablation studies}
\label{sec: Additional ablation study}
We conduct additional ablation studies, including the effect of different sampling policies and different maximum lengths of generated sequences.

\subsection{different sampling policies} In our default settings,  VLMs generate the sequences using nucleus sampling method \citep{holtzman2019curious} with $p=0.9$ and temperature $t=1$. Besides, we present the results of  three  other
sampling policies, including greedy search, beam search with a beam width of 5, and top-k sampling with $k=10$. As depicted in  Table \ref{tab:different sampling policies}, our verbose images can induce VLMs to generate the longest sequences across various sampling policies, demonstrating that our verbose images are not sensitive to the generation sampling policies.

\subsection{different maximum lengths of generated sequences}
Table \ref{tab:different maximum length} presents the results of five categories of visual images under varying maximum lengths of generated sequences.
It shows that as the maximum length of generated sequences of VLMs increases, the length of generated sequences becomes longer, leading to higher energy consumption and longer latency time. 
Furthermore, our verbose images can consistently outperform the other four methods, which confirms the superiority of our verbose images.

\section{Grid search}
\label{sec: Grid search}
We conduct the experiments of grid search \citep{gao2023backdoor} for different parameters of loss weights and different momentum values.

\subsection{different parameters of loss weights}
We set parameters of loss weights as $a_1=10$, $b_1=-20$, $a_2=0$, $b_2=0$, $a_3=0.5$, and $b_3=1$ during the optimization of our verbose images. These parameters are determined through the grid search. The results of grid search are shown in Table \ref{tab:grid search 1}, Table \ref{tab:grid search 2}, Table \ref{tab:grid search 3}, Table \ref{tab:grid search 4}, Table \ref{tab:grid search 5}, and Table \ref{tab:grid search 6}, which demonstrates that our verbose images with $a_1=10$, $b_1=-20$, $a_2=0$, $b_2=0$, $a_3=0.5$, and $b_3=1$ can induce VLMs to generate the longest sequences.

\subsection{different momentum values}
We show the results of our verbose images in different momentum values $m=\{0, 0.3, 0.6, 0.9\}$ in Table \ref{tab:momentum optimization}. These results demonstrate that it is necessary to adopt the addition of momentum during the optimization.

\section{Visualization}
\label{sec: Visualization}
\begin{figure*}[t] \centering   
\begin{minipage}{\textwidth}
    \includegraphics[width=\columnwidth]{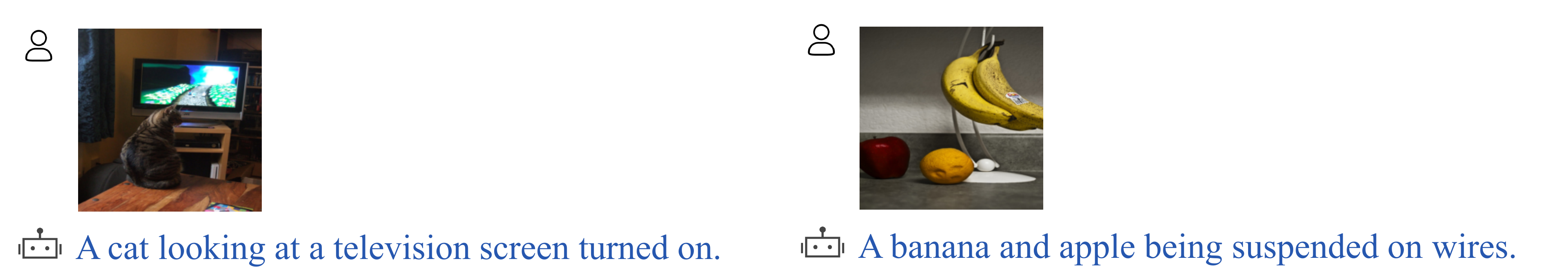}  
\end{minipage}
\begin{minipage}{\textwidth}
    \includegraphics[width=0.995
    \columnwidth]{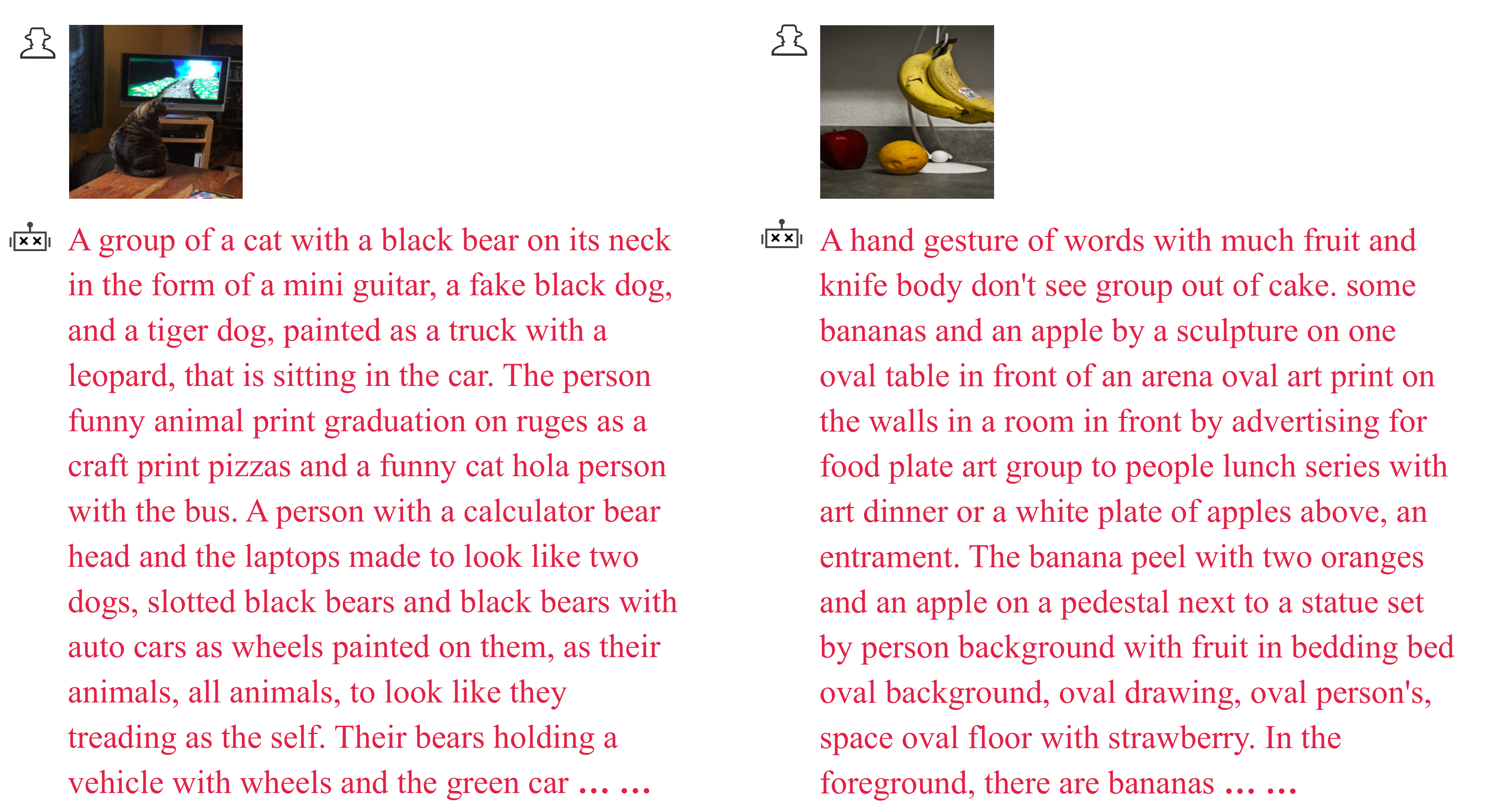}  
\end{minipage}
\vspace{-1em}
\caption{A visualization example for the original images and our verbose counterpart against BLIP.} 
\label{blip appendix}
\end{figure*}

\begin{figure*}[t] \centering   
\begin{minipage}{\textwidth}
    \includegraphics[width=\columnwidth]{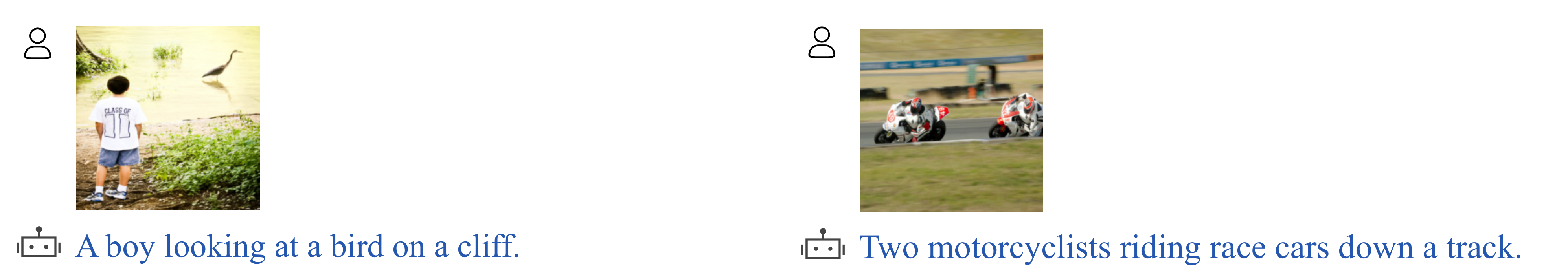}  
\end{minipage}
\begin{minipage}{\textwidth}
    \includegraphics[width=\columnwidth]{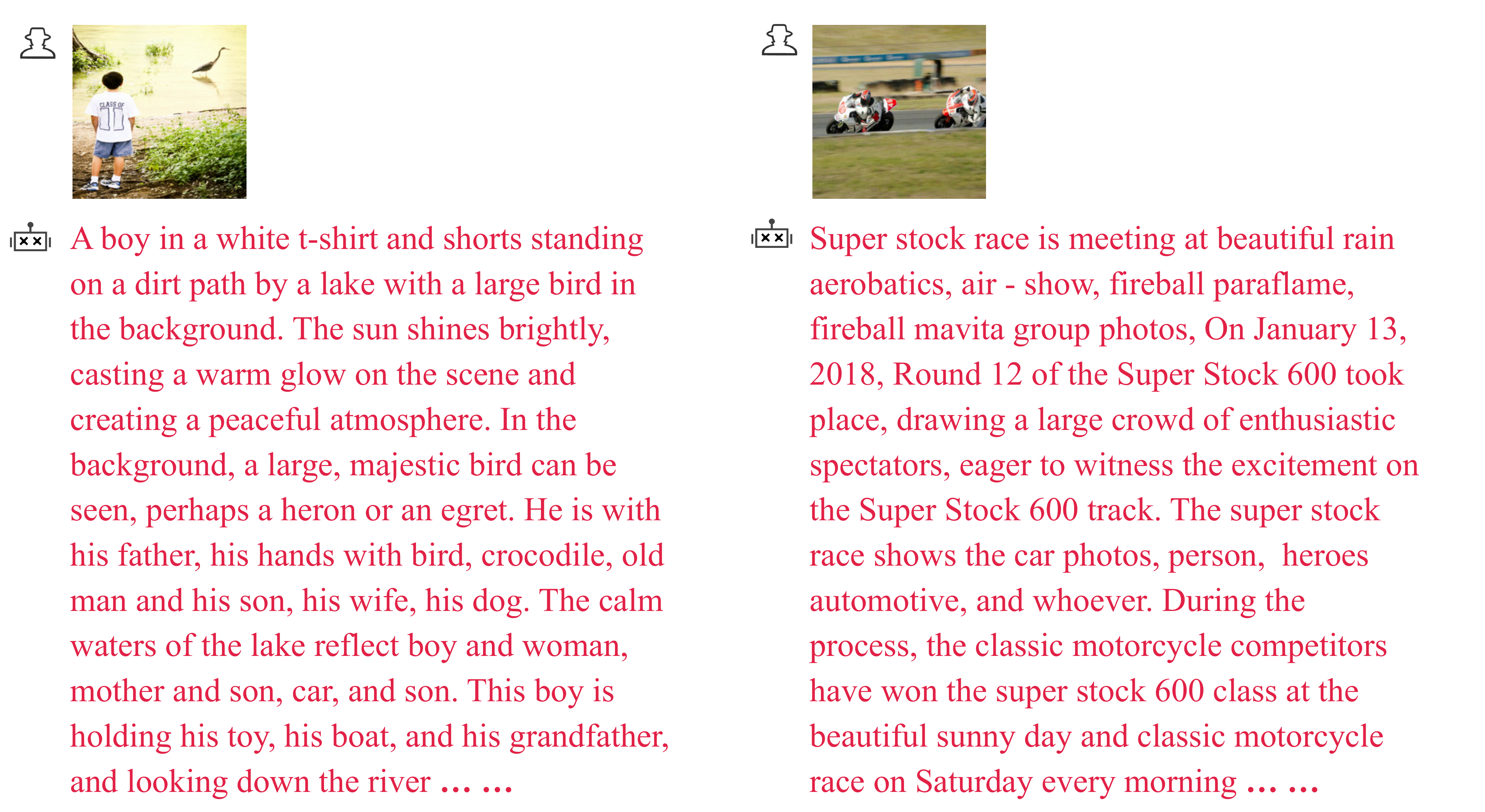}  
\end{minipage}
\vspace{-1em}
\caption{A visualization example for the original images and our verbose counterpart against BLIP-2.} 
\label{blip2 appendix}
\end{figure*}

\begin{figure*}[t] \centering   
\begin{minipage}{\textwidth}
    \includegraphics[width=\columnwidth]{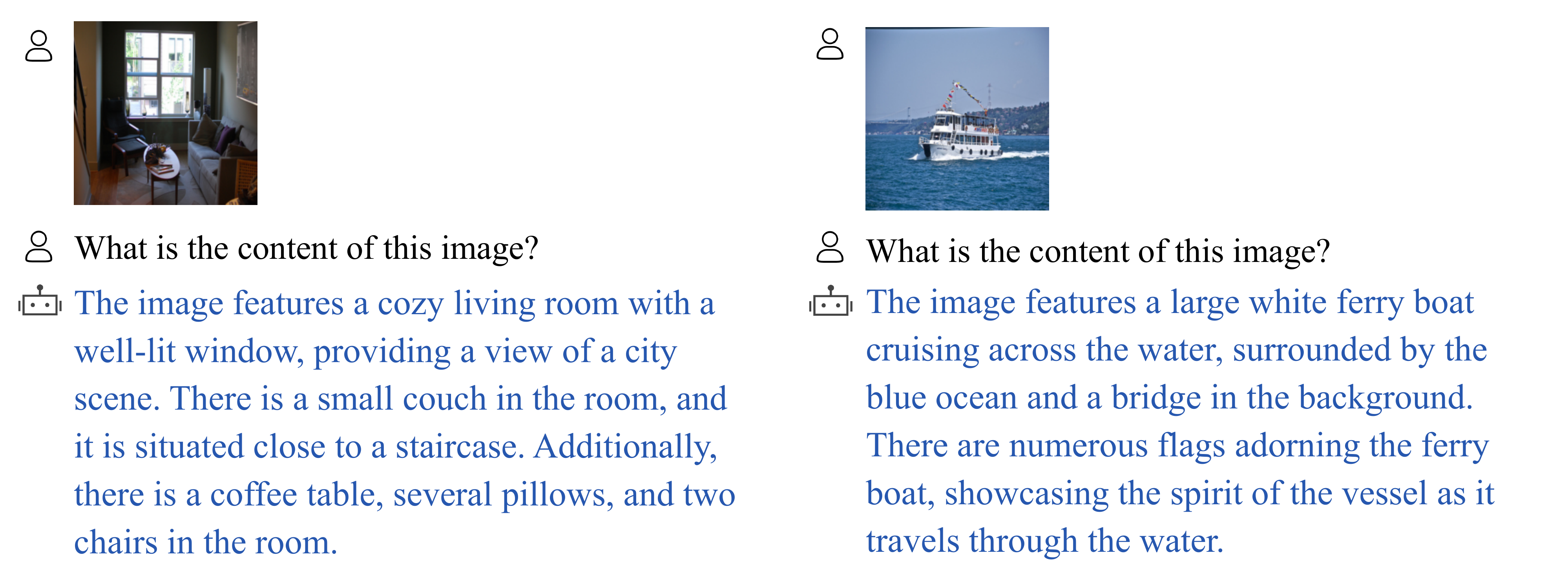}  
\end{minipage}
\begin{minipage}{\textwidth}
    \includegraphics[width=\columnwidth]{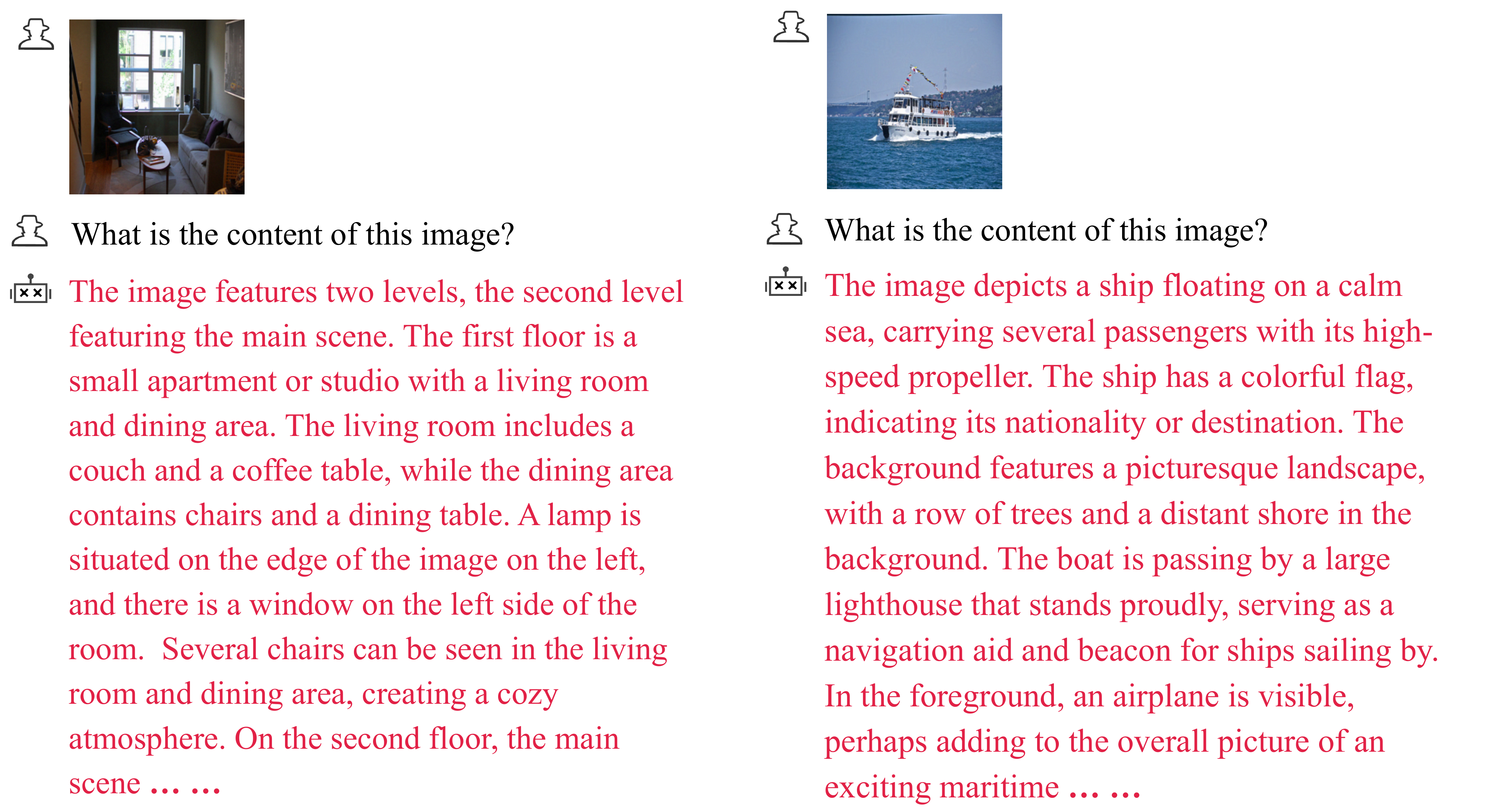}  
\end{minipage}
\vspace{-1em}
\caption{A visualization example for the original images and our verbose counterpart against InstructBLIP.} 
\label{instr appendix}
\end{figure*}

\begin{figure*}[t] \centering   
\begin{minipage}{\textwidth}
    \includegraphics[width=\columnwidth]{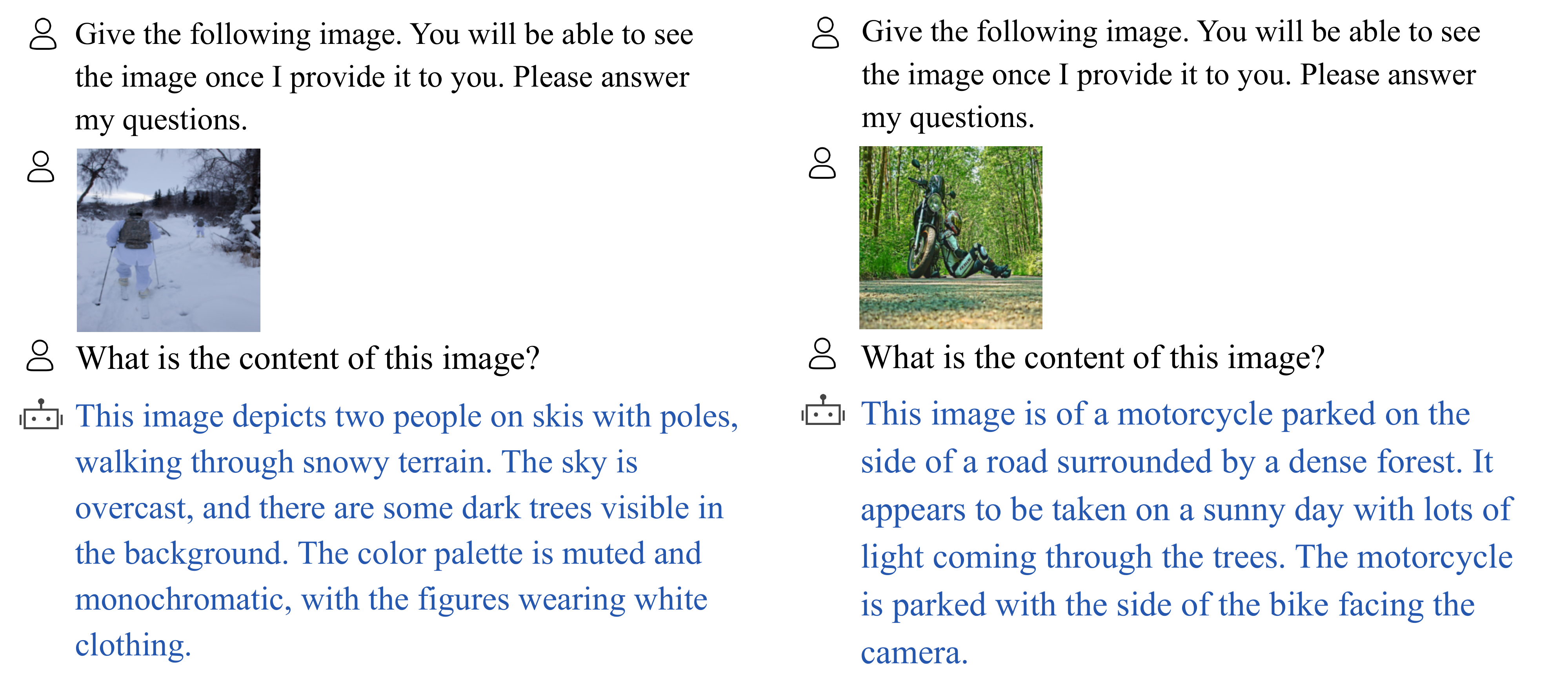}  
\end{minipage}
\begin{minipage}{\textwidth}
    \includegraphics[width=\columnwidth]{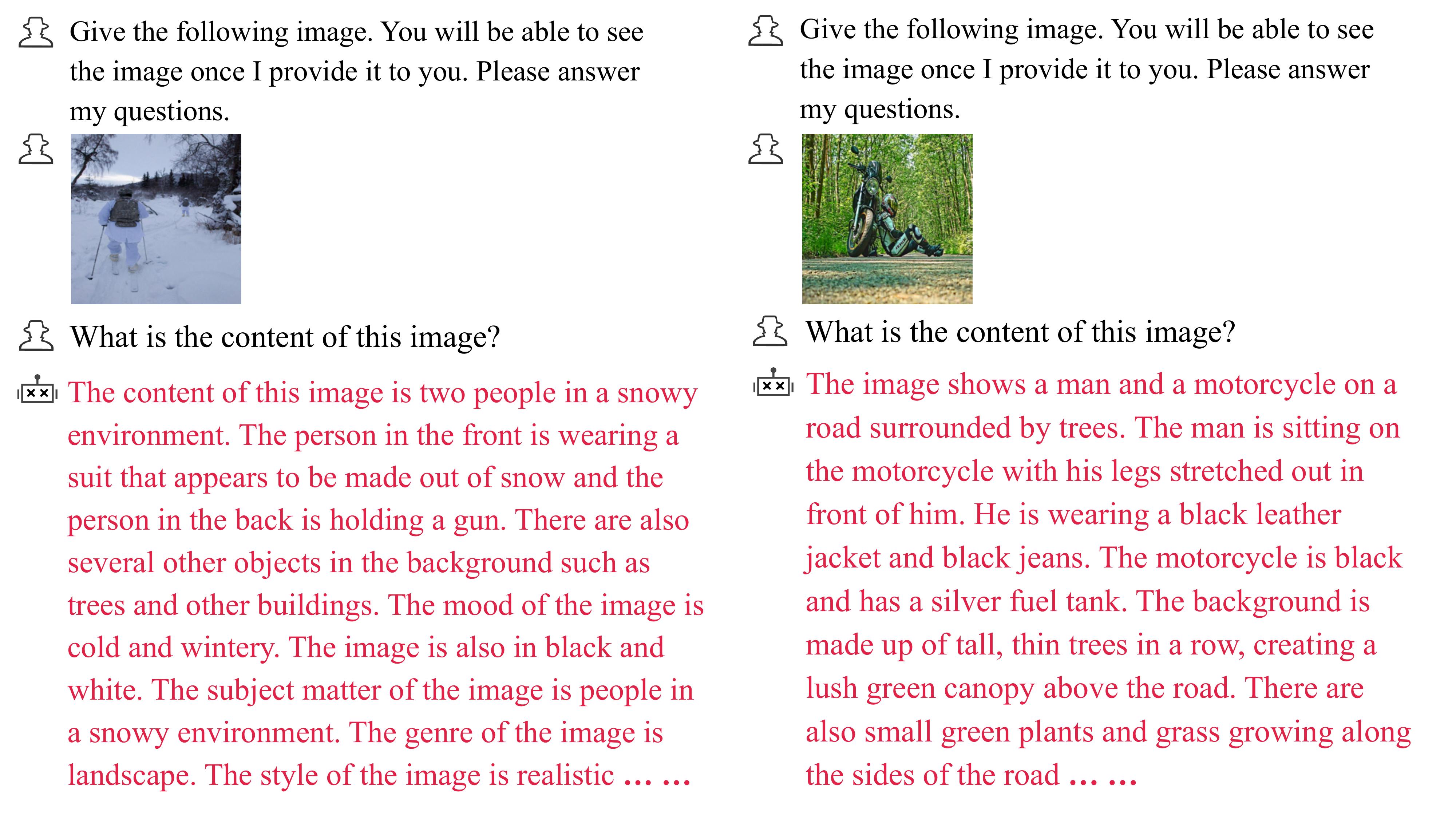}  
\end{minipage}
\vspace{-1em}
\caption{A visualization example for the original images and our verbose counterpart against MiniGPT-4.} 
\label{gpt appendix}
\end{figure*}

We visualize the examples of the original images and our verbose images against BLIP, BLIP-2, InstructBLIP, and MiniGPT-4 in Fig. \ref{blip appendix}, Fig. \ref{blip2 appendix}, Fig. \ref{instr appendix}, and Fig. \ref{gpt appendix}. It can be observed that VLMs with more advanced LLMs (\textit{e.g.}, Vicuna-7B) generate more fluent, smooth, and logical content when encountering our verbose images.

\begin{table*}[]
\begin{minipage}{\textwidth}
\caption{The length of generated sequences, energy consumption (J), and latency time (s) against BLIP-2 for grid search of $a_1$.}
\label{tab:grid search 1}
\centering
\small
\setlength\tabcolsep{7.9pt}{
\begin{tabular}{@{}cccccc|ccc|ccc@{}}
\toprule
\multirow{2}{*}{$a_1$} & \multicolumn{1}{c}{\multirow{2}{*}{$b_1$}} & \multicolumn{1}{c}{\multirow{2}{*}{$a_2$}} & \multicolumn{1}{c}{\multirow{2}{*}{$b_2$}} & \multicolumn{1}{c}{\multirow{2}{*}{$a_3$}} & \multicolumn{1}{c|}{\multirow{2}{*}{$b_3$}} & \multicolumn{3}{c|}{MS-COCO} & \multicolumn{3}{c}{ImageNet} \\
 & & & & & & Length & Latency & Energy & Length & Latency & Energy  \\
\midrule 
1 & -20 & 0 & 0 & 0.5 & 1 & 215.91 & 7.36 & 310.36 & 233.67 & 8.83 & 375.31 \\
\textbf{10} & \textbf{-20} & \textbf{0} & \textbf{0} & \textbf{0.5} & \textbf{1} & \textbf{226.72} &	\textbf{7.97} & \textbf{321.59} & \textbf{250.72} & \textbf{10.26} & \textbf{398.58} \\
100 & -20 & 0 & 0 & 0.5 & 1 & 159.34 & 5.77 & 207.95 & 162.75 & 6.88 & 232.95 \\
\bottomrule
\end{tabular}}
\end{minipage}

\begin{minipage}{\textwidth}
\caption{The length of generated sequences, energy consumption (J), and latency time (s) against BLIP-2 for grid search of $b_1$.}
\label{tab:grid search 2}
\centering
\small
\setlength\tabcolsep{7.9pt}{
\begin{tabular}{@{}cccccc|ccc|ccc@{}}
\toprule
\multirow{2}{*}{$a_1$} & \multicolumn{1}{c}{\multirow{2}{*}{$b_1$}} & \multicolumn{1}{c}{\multirow{2}{*}{$a_2$}} & \multicolumn{1}{c}{\multirow{2}{*}{$b_2$}} & \multicolumn{1}{c}{\multirow{2}{*}{$a_3$}} & \multicolumn{1}{c|}{\multirow{2}{*}{$b_3$}} & \multicolumn{3}{c|}{MS-COCO} & \multicolumn{3}{c}{ImageNet} \\
 & & & & & & Length & Latency & Energy & Length & Latency & Energy  \\
\midrule 
10 & -2 & 0 & 0 & 0.5 & 1 & 212.99 & 7.39 & 307.97 & 212.14 & 7.56 & 304.26  \\
\textbf{10} & \textbf{-20} & \textbf{0} & \textbf{0} & \textbf{0.5} & \textbf{1} & \textbf{226.72} &	\textbf{7.97} & \textbf{321.59} & \textbf{250.72} & \textbf{10.26} & \textbf{398.58} \\
10 & -200 & 0 & 0 & 0.5 & 1 & 195.23 & 6.74 & 253.56 & 195.38 & 6.32 & 266.37 \\
\bottomrule
\end{tabular}}
\end{minipage}

\begin{minipage}{\textwidth}
\caption{The length of generated sequences, energy consumption (J), and latency time (s) against BLIP-2 for grid search of $a_2$.}
\label{tab:grid search 3}
\centering
\small
\setlength\tabcolsep{7.9pt}{
\begin{tabular}{@{}cccccc|ccc|ccc@{}}
\toprule
\multirow{2}{*}{$a_1$} & \multicolumn{1}{c}{\multirow{2}{*}{$b_1$}} & \multicolumn{1}{c}{\multirow{2}{*}{$a_2$}} & \multicolumn{1}{c}{\multirow{2}{*}{$b_2$}} & \multicolumn{1}{c}{\multirow{2}{*}{$a_3$}} & \multicolumn{1}{c|}{\multirow{2}{*}{$b_3$}} & \multicolumn{3}{c|}{MS-COCO} & \multicolumn{3}{c}{ImageNet} \\
 & & & & & & Length & Latency & Energy & Length & Latency & Energy  \\
\midrule 
10 & -20 & 0.1 & 0 & 0.5 & 1 & 177.59 & 7.06 & 270.62 & 213.63 & 8.18 & 297.52  \\
\textbf{10} & \textbf{-20} & \textbf{0} & \textbf{0} & \textbf{0.5} & \textbf{1} & \textbf{226.72} &	\textbf{7.97} & \textbf{321.59} & \textbf{250.72} & \textbf{10.26} & \textbf{398.58} \\
10 & -20 & 1 & 0 & 0.5 & 1 & 196.17 & 6.87 & 290.8 & 227.38 & 10.06 & 357.64 \\
\bottomrule
\end{tabular}}
\end{minipage}

\begin{minipage}{\textwidth}
\caption{The length of generated sequences, energy consumption (J), and latency time (s) against BLIP-2 for grid search of $b_2$.}
\label{tab:grid search 4}
\centering
\small
\setlength\tabcolsep{7.9pt}{
\begin{tabular}{@{}cccccc|ccc|ccc@{}}
\toprule
\multirow{2}{*}{$a_1$} & \multicolumn{1}{c}{\multirow{2}{*}{$b_1$}} & \multicolumn{1}{c}{\multirow{2}{*}{$a_2$}} & \multicolumn{1}{c}{\multirow{2}{*}{$b_2$}} & \multicolumn{1}{c}{\multirow{2}{*}{$a_3$}} & \multicolumn{1}{c|}{\multirow{2}{*}{$b_3$}} & \multicolumn{3}{c|}{MS-COCO} & \multicolumn{3}{c}{ImageNet} \\
 & & & & & & Length & Latency & Energy & Length & Latency & Energy  \\
\midrule 
10 & -20 & 0 & 0.1 & 0.5 & 1 & 132.13 & 5.54 & 206.47 & 160.53 & 6.51 & 227.87  \\
\textbf{10} & \textbf{-20} & \textbf{0} & \textbf{0} & \textbf{0.5} & \textbf{1} & \textbf{226.72} &	\textbf{7.97} & \textbf{321.59} & \textbf{250.72} & \textbf{10.26} & \textbf{398.58} \\
10 & -20 & 0 & 1 & 0.5 & 1 & 203.61 & 7.08 & 286.32 & 224.06 & 9.36 & 345.61 \\
\bottomrule
\end{tabular}}
\end{minipage}

\begin{minipage}{\textwidth}
\caption{The length of generated sequences, energy consumption (J), and latency time (s) against BLIP-2 for grid search of $a_3$.}
\label{tab:grid search 5}
\centering
\small
\setlength\tabcolsep{7.9pt}{
\begin{tabular}{@{}cccccc|ccc|ccc@{}}
\toprule
\multirow{2}{*}{$a_1$} & \multicolumn{1}{c}{\multirow{2}{*}{$b_1$}} & \multicolumn{1}{c}{\multirow{2}{*}{$a_2$}} & \multicolumn{1}{c}{\multirow{2}{*}{$b_2$}} & \multicolumn{1}{c}{\multirow{2}{*}{$a_3$}} & \multicolumn{1}{c|}{\multirow{2}{*}{$b_3$}} & \multicolumn{3}{c|}{MS-COCO} & \multicolumn{3}{c}{ImageNet} \\
 & & & & & & Length & Latency & Energy & Length & Latency & Energy  \\
\midrule 
10 & -20 & 0 & 0 & 0.05 & 1 & 209.43 & 7.18 & 292.43 & 217.03 & 8.62 & 303.73  \\
\textbf{10} & \textbf{-20} & \textbf{0} & \textbf{0} & \textbf{0.5} & \textbf{1} & \textbf{226.72} &	\textbf{7.97} & \textbf{321.59} & \textbf{250.72} & \textbf{10.26} & \textbf{398.58} \\
10 & -20 & 0 & 0 & 5 & 1 & 199.62 & 7.17 & 283.76 & 226.52 & 8.06 & 335.48 \\
\bottomrule
\end{tabular}}
\end{minipage}

\begin{minipage}{\textwidth}
\caption{The length of generated sequences, energy consumption (J), and latency time (s) against BLIP-2 for grid search of $b_3$.}
\label{tab:grid search 6}
\centering
\small
\setlength\tabcolsep{7.9pt}{
\begin{tabular}{@{}cccccc|ccc|ccc@{}}
\toprule
\multirow{2}{*}{$a_1$} & \multicolumn{1}{c}{\multirow{2}{*}{$b_1$}} & \multicolumn{1}{c}{\multirow{2}{*}{$a_2$}} & \multicolumn{1}{c}{\multirow{2}{*}{$b_2$}} & \multicolumn{1}{c}{\multirow{2}{*}{$a_3$}} & \multicolumn{1}{c|}{\multirow{2}{*}{$b_3$}} & \multicolumn{3}{c|}{MS-COCO} & \multicolumn{3}{c}{ImageNet} \\
 & & & & & & Length & Latency & Energy & Length & Latency & Energy  \\
\midrule 
10 & -20 & 0 & 0 & 0.5 & 0.1 & 208.48 & 7.16 & 290.92 & 236.06 & 9.97 & 380.23  \\
\textbf{10} & \textbf{-20} & \textbf{0} & \textbf{0} & \textbf{0.5} & \textbf{1} & \textbf{226.72} &	\textbf{7.97} & \textbf{321.59} & \textbf{250.72} & \textbf{10.26} & \textbf{398.58} \\
10 & -20 & 0 & 0 & 0.5 & 10 & 186.68 & 7.05 & 282.61 & 222.03 & 9.03 & 340.55 \\
\bottomrule
\end{tabular}}
\end{minipage}

\begin{minipage}{\textwidth}
\caption{The length of generated sequences, energy consumption (J), and latency time (s) against BLIP-2 for grid search of the momentum $m$.}
\label{tab:momentum optimization}
\centering
\small
\setlength\tabcolsep{18pt}{
\begin{tabular}{@{}c|ccc|ccc@{}}
\toprule
\multirow{2}{*}{$m$} & \multicolumn{3}{c|}{MS-COCO} & \multicolumn{3}{c}{ImageNet} \\
 & Length & Latency & Energy & Length & Latency & Energy  \\
\midrule 
0 & 199.92 & 7.02 & 292.55 & 231.03 & 7.88 & 318.34 \\
0.3 & 201.83 & 7.21 & 301.70 & 237.85 & 9.33 & 330.83 \\
0.6 & 218.69 & 7.69 & 312.77 &  240.27 & 9.85 & 376.59 \\
\textbf{0.9} & \textbf{226.72} &	\textbf{7.97} & \textbf{321.59} & \textbf{250.72} & \textbf{10.26} & \textbf{398.58}  \\
\bottomrule
\end{tabular}}
\end{minipage}
\end{table*}

\clearpage

\end{document}